\documentclass[times, review, 10pt]{elsarticle}

\usepackage{setspace}
\usepackage[numbers]{natbib}
\usepackage{amsmath,amsfonts}
\usepackage{color}
\usepackage{amssymb}
\usepackage{amsmath}

\usepackage{booktabs}
\usepackage{multirow}
\usepackage{amssymb}
\usepackage{array}
\usepackage{paralist}       
\usepackage{amsmath} 
\usepackage{url}
\usepackage{stfloats}
\usepackage{subcaption}
 
\def\eg{\emph{e.g.}} 

\def\vs{\emph{v.s.}} 
\def\ie{\emph{i.e.}}

\journal{Elsevier}

\begin{document}
\definecolor{gray}{rgb}{0.5, 0.5, 0.5}  
\begin{frontmatter}



\title{RETTA: Retrieval-Enhanced Test-Time Adaptation for Zero-Shot Video Captioning}






\author[1]{Yunchuan Ma}
\ead{mayunchuan23@mails.ucas.ac.cn}
\author[1]{Laiyun Qing \corref{cor1}}
\author[1]{Guorong Li}
\author[2]{Yuankai Qi}
\author[2]{Amin Beheshti}
\author[2]{Quan Z. Sheng}
\author[1]{Qingming Huang}

\cortext[cor1]{Corresponding author}

\address[1]{University of Chinese Academy of Science, Beijing,100190, China}
\address[2]{Macquarie University}


\begin{abstract}
 Despite the significant progress of fully-supervised video captioning, zero-shot methods remain much less explored.
 %
In this paper, we propose a novel zero-shot video captioning framework named \textbf{R}etrieval-\textbf{E}nhanced \textbf{T}est-\textbf{T}ime \textbf{A}daptation (RETTA), which takes advantage of existing pre-trained large-scale vision and language models to directly generate captions with 
test-time adaptation.
%
Specifically, we bridge video and text using four key models: a general video-text retrieval model XCLIP, a general image-text matching model CLIP, a text alignment model AnglE, and a text generation model GPT-2, due to their source-code availability.
The main 
challenge 
is how to enable the text generation model to be sufficiently aware of the content in a given video so as to generate corresponding captions.
To address this problem, we propose using learnable tokens as a communication medium among these four frozen models GPT-2, XCLIP, CLIP, and AnglE.
Different from the conventional way that trains these tokens with training data, we propose to learn these tokens with soft targets of the inference data under several carefully crafted loss functions, which enable the tokens to absorb video information catered for GPT-2.
This adaptation requires only a few iterations (\eg, 16) and does not require ground truth data.
  %
  %
  %
Extensive experimental on MSR-VTT, MSVD, and VATEX, show absolute 
5.1\%$\sim$32.4\% improvements in CIDEr scores compared to 
several state-of-the-art zero-shot video captioning methods.
\end{abstract}

\end{frontmatter}


\section{Introduction}
\label{sec:intro}




\begin{figure}[!t]
\centering
\includegraphics[width=0.7\linewidth]{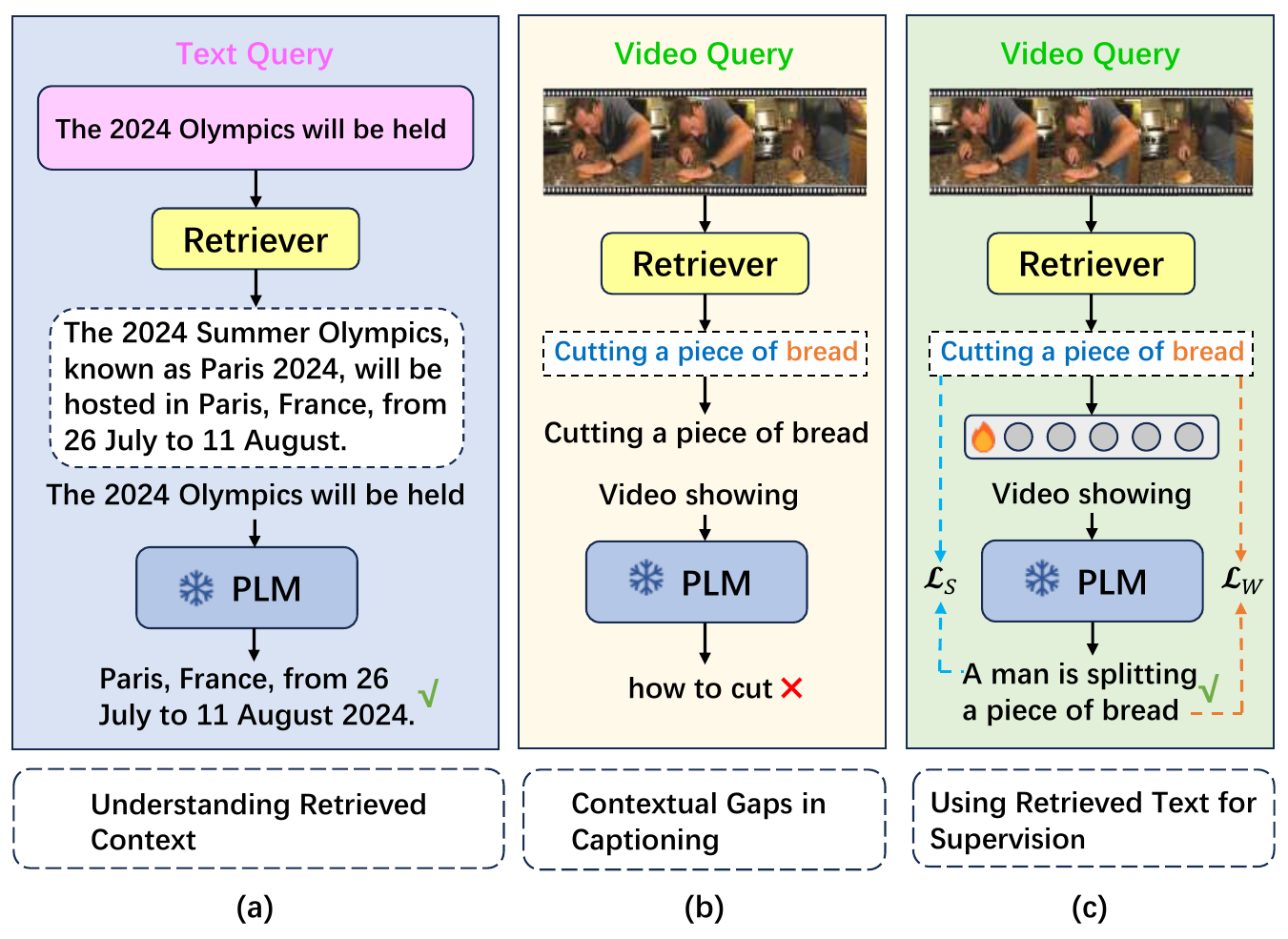}
\caption{Traditional retrieval-augmented-generation paradigm performs well on pure natural language tasks, such as text-based question answering. However, its performance sucks when directly apply to multi-modality tasks, such as video captioning.
}
\label{fig:insight}
\end{figure}

Video captioning, aiming to automatically describe the video content in one sentence, is one of the traditional vision-language tasks. 
It plays a crucial role in numerous applications, such as video title generation~\cite{video-title-generation}, blind assistance~\cite{wenlan}, and video search~\cite{Search-oriented}.
\textcolor{black}{
However, current video captioning research~\cite{pr3, pr4, pr5} typically requires full supervision, with at least 10 reference captions per video, leading to high annotation costs.}
\textcolor{black}{As a practical solution, zero-shot methods reduce the dependence on costly human annotations~\cite{MultiCapCLIP,DeCap}.
This makes them particularly suitable for scenarios where labeled data is scarce.
}

Recently, with the development of pre-trained foundation models in vision and language areas, their combination has shown outstanding zero-shot 
learning capabilities~\cite{MeaCap,MERCap}.
%
For example, for the text-based question-answering task, pre-trained retrieval models have been integrated with Pre-trained Large language Models (PLMs) to improve the quality of generated answers.
An example is shown in Figure~\ref{fig:insight}~(a), where the retriever takes the question ``The 2024 Olympics will be held'' as a query to retrieve relevant knowledge ``The 2024 Summer Olympics, known as Paris 2024, will be hosted in Paris, France, from 26 July to 11 August.''. By concatenating it with the question as a prompt for the model, we get the output   ``Paris, France, from 26 July to 11 August 2024''.

However, a simple application of this paradigm does not work well for multi-modal visual description tasks~\cite{basketball1,basketball2} (\eg, video captioning). 
For example, as shown in Figure~\ref{fig:insight}~(b), the video query retrieves the related sentence ``Cutting a piece of bread''; then together with the hard prompt ``Video showing'' serving as the input of PLM,  the PLM outputs ``how to cut'',
which does not meet the expectation of video captioning. 
The main reasons lie in two aspects:
(1) Weak semantic connection: The semantic connection between the prompt ``video showing'' and the retrieved text is not as strong as that between questions and answers in QA tasks.
\textcolor{black}{
This is because traditional RAG methods~\cite{icl,iclqa} typically use the question as a query to retrieve relevant textual information as a prefix, which effectively guides the model to generate accurate responses.
In contrast, in video captioning, when the video itself is used as a query~\cite{MV-Adapter,pr6}, the retrieved text is often weakly related to generic prompts like ``video showing'', thereby hindering PLMs from generating the expected outputs.}
(2) Improper prompts:
\textcolor{black}{Prompt engineering itself is a challenging task~\cite{Improper_prompts1,Improper_prompts2}}, letting alone to additionally take diverse retrieved text into consideration.

To address the above-mentioned problems, we propose to leverage learnable tokens as a connection medium among different foundation models. 
\textcolor{black}{Inspired by test-time prompt tuning approaches~\cite{tpt, vpa}, the learnable tokens are updated during inference using gradient signals from the task loss.
Their role is to deliver video-specific information to the PLMs and specify the expected form of output.}
%
%
Moreover, we expect the combination of these pre-trained models is able to quickly adapt to the task, preferably without the need of training.
To meet the quick adaptation requirement, we propose to update these tokens during inference but with soft targets~\cite{rcg, pr8}, which gets rid of the conventional training of soft prompts. 
\textcolor{black}{These tokens thus serve as a bridge between different frozen models, making them a practical and effective connection strategy in our setting.}
Our experiments show that the whole framework can well adapt to the video captioning task after just a few of iterations (we use 16 iterations).
To meet the former content awareness and output type requirements, we design several loss functions aligning the generated text and soft targets.
\textcolor{black}{
Specifically, the soft targets are calculated by retrieved text and candidate sentences to guide the choice of candidate words at each time step.
This is because soft targets~\cite{rcg, pr8} can help the model consider multiple potential tokens in text generation, boosting the diversity of the generated results and improving generalization.
}
An example is shown in 
Figure~\ref{fig:insight}(c), where our method leads to better video captions.
%
%

In summary, the main contributions of this paper are as follows:
\begin{compactitem}
\item To the best of our knowledge, we propose the first fast adaptation framework that combines a pre-trained video-text retrieval model, an image-text matching model, {a text alignment model}, and a language generation model for video captioning. 
\item We design two loss functions that effectively convey retrieved information to PLMs to generate captions closely related to video content.
%
\item Extensive experimental results on three widely-used datasets show that our method outperforms several state-of-the-art zero-shot {video captioning approaches with 
absolute 
5.1\%$\sim$32.4\%}  improvement in terms of the main metric CIDEr.

\end{compactitem}

\section{Related work}

\noindent\textbf{Video Captioning}
Recently, a series of full-supervision video captioning research~\cite{pr1, pr2} have made significant progress.
However, fewer efforts have been made on the video captioning task under the zero-shot setting~\cite{ZeroCap,ept}.
ZeroCap~\cite{ZeroCap} uses a visual-language model to guide a language model to describe the given image via the designed CLIP-loss.
EPT~\cite{ept} further extends it to video captioning using the proposed CLIP-loss to optimize the pseudo-tokens.
To improve inference speed, MAGIC~\cite{MAGIC} optimizes the decoding strategy of language models and employs CLIP scores to adjust PLM logits for better alignment with image correspondence.
%
DeCap~\cite{DeCap} and MultiCapCLIP~\cite{MultiCapCLIP} also leverage the visual-language alignment ability of CLIP for zero-shot video captioning. They first utilize text-only data to train a decoder from scratch and then input video to generate captions at the inference stage.
%
%
Although the above-mentioned methods improve zero-shot caption quality, they still face the challenge of understanding video content.
\textcolor{black}{In addition, using CLIP in open-world scenarios may lead to degraded performance due to attention misalignment and bag-of-words behavior~\cite{bag-of-words}. These issues introduce biases in multi-object scenes, where CLIP tends to focus on larger objects and overlook text order. This suggests that relying solely on CLIP may be insufficient for tackling open-world tasks such as zero-shot video captioning.}
To tackle these 
problems, we propose a retrieval-augmented video captioning framework that retrieves video-related knowledge to deepen the understanding of the video.

\noindent\textbf{Video-Text Retrieval}
Video-text retrieval (VTR) aims to find the most relevant text based on 
a 
video query.
To retrieve the correct video candidates given 
a 
text query, recent works~\cite{pr6,pr7} focus on the cross-modal alignment between the related videos and text samples.
%
%
To facilitate this alignment, some works leverage pre-trained models such as CLIP as the backbone to obtain rich cross-modal representations.
For example, Clip4clip~\cite{clip4clip} transfers the image-text knowledge of the CLIP model to the VTR task in a coarse manner.
To better adapt CLIP to the video domain, X-CLIP~\cite{xclip} extends CLIP for general video-language understanding using the Kinetics-600 dataset. 
%
In this work, we adopt~\cite{xclip} as our retriever
as its training data has no overlap with video captioning datasets.
 
\noindent\textbf{Retrieval-Augmented Generation}
Retrieval-Augmented Generation (RAG) augments the generation ability of language models with retrieved knowledge.
Generally, the knowledge retrieved is used as a prefix input to the pre-trained language model (PLM), prompting PLM to generate more accurate answers.
REALM~\cite{realm} employs a latent knowledge retriever to boost language model pre-training, which allows capturing more real-world knowledge in an interpretable manner. 
ORQA~\cite{orqa} pre-trains the retriever with an unsupervised inverse cloze task and learns to retrieve evidence from an open corpus. 
DPR~\cite{DPR} implements a dense embedding retriever to better select candidate contexts using only pairs of questions and answers, without additional pretraining.
RCG~\cite{rcg} plugs a cross-modal retriever to provide useful hints and employs a dynamic copy-mechanism generator for better generation.
Recently, with the development of large language models (LLMs), a series of RAG-related LLM 
approaches
are proposed. RECOMP~\cite{RECOMP} uses a summarization model to compress retrieved documents into a textual summary before prompting the LLM to generate the output. 
Chen~\textit{et al.} establish the first benchmark to evaluate robustness-related capabilities for retrieval augmented generation of LLMs.
Asai~\textit{et al.} further introduce a new framework SELF-RAG, which trains an arbitrary LM to generate text with reflection tokens by integrating them as the next token prediction from the expanded model vocabulary.

Although RAG has been applied to fully-supervised video captioning, there is no related work on zero-shot video captioning tasks, to the best of our knowledge.
It faces the challenge of how retrieved sentences can guide the LLMs in generating appropriate video descriptions.
To solve this problem, we design two loss functions to optimize learnable tokens via aligning LLMs' output and retrieval results. 


\begin{figure*}[!t]
\centering
\includegraphics[width=0.7\linewidth]{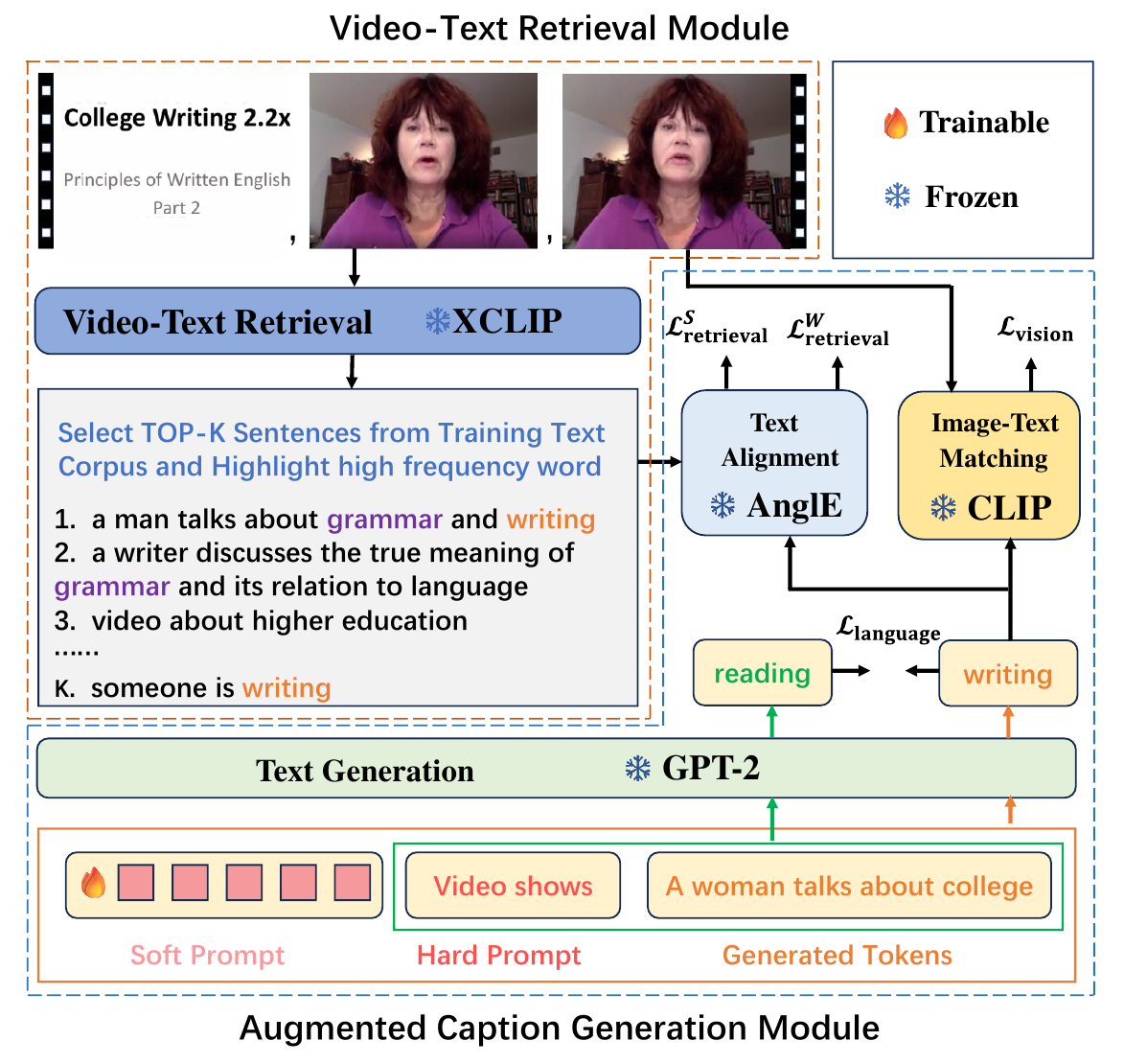}
\caption{The workflow of the proposed method, which consists of frozen pre-trained foundation vision and language models as well as trainable tokens. Unlike conventional soft prompt training, we update these tokens directly during inference with soft target. In this way, the frozen foundation models can quickly adapt to the video captioning task for zero-shot application.
\textcolor{black}{
$\mathcal{L}_{language}$ is computed between token probability distributions generated with (\ie~writing) and without (\ie~reading) the trainable soft prompt.}
$\mathcal{L}_{vision}$ calculates a matching score between the current generated text and visual information leveraging CLIP based on keyframes.
$\mathcal{L}_{retrieval}^S$ measures a matching score between the current generated text and retrieved sentences using AnglE, and $\mathcal{L}_{retrieval}^W$ focuses on the high-frequency words of them. 
The above optimizations take place at each step of the autoregressive process.
}
\label{fig:method}
\end{figure*}

\section{Method}
%

As shown in Figure~\ref{fig:method}, our framework contains four pre-trained frozen models: XCLIP~\cite{xclip} and CLIP~\cite{clip} provide video-level and frame-level video understanding; GPT-2~\cite{gpt2} provide the text generation capability; AnglE~\cite{angle} is used to calculate text-text similarity.
According to their functions for the video captioning task, these pre-trained models can be grouped into a video-text retrieval module and an augmented caption generation module.
\textcolor{black}{
The video-text retrieval module retrieves relevant sentences from the corpus and highlights key high-frequency words. These elements serve as auxiliary semantic cues for the caption generation module, enabling it to generate more accurate and contextually appropriate captions.
}
Specifically, XCLIP serves as the video-text retrieval module, providing a basic understanding of the video. 
It retrieves video content-related sentences {from training text corpus}, and we highlight its high-frequency words, as detailed in Section~\ref{sec:retriever}.
The remaining pre-trained models compose the augment caption generation module,
%
%
which utilizes several test-time learnable tokens to connect all these pre-trained models, absorb video information, and serve as proper prompts for text generation as detailed in Section~\ref{sec:captiongenerator}.
In Section~\ref{sec: inference}, we provide more details of the inference process.

\subsection{Video-Text Retrieval}\label{sec:retriever}
The video-text retrieval module adopts a multi-modal retriever XCLIP to retrieve video-related descriptions from a text corpus.
XCLIP is pre-trained on the Kinetics-600 dataset and has the zero-shot ability for video classification and video-text retrieval tasks.
%
Traditional retrievers consistently fetch the top-$k$ ranked sentences. However, this retrieval strategy faces two challenges for video captioning. First, choosing the hyper-parameter $k$ is challenging since the optimal $k$ varies across different videos. Second, retrieving a fixed number of $k$ sentences may introduce noise, as not all of these sentences are closely related to the video.

To mitigate the aforementioned issues, we propose to additionally sample top-$L$ high-frequency words (limited to nouns and verbs) from the retrieved sentences, which reduces diversity of sentences.
We detail how retrieved sentences and sampled high-frequency words are utilized to guide caption generation in the next section.
\subsection{Augmented Caption Generation}
\label{sec:captiongenerator}
With the help of the video-text retrieval module, the augmented caption generation module uses the retrieved sentences and high-frequency words to improve the quality of the generated captions. 
This module contains three main components: the large language model GPT-2, the image-text matching model CLIP, and the text alignment model AnglE.
The out-of-box GPT-2 is pre-trained on the large dataset WebText collected from a wide range of text data online.
It is composed of several Transformer layers, which can cache the previous key and value tokens to infer the next token autoregressively.
Therefore, GPT-2 is capable of 
generating text based on preceding content.
The CLIP model is pre-trained on 400 million image-text pairs, which plays the vision-language alignment role to map the image and sentence into a unified semantic space.
AnglE is a state-of-the-art model for calculating semantic textual similarity, which is able to be aware of differences among texts.

As shown in Figure~\ref{fig:method}, all model parameters are frozen, and only the learnable soft prompts
are optimized during the inference process with soft targets to guide GPT-2 to generate text in line with the content of the given video.
Take the $i-{th}$ step as an example, the text generation can be described as:
\begin{equation}
    w_{i+1} = \operatorname{PLM}(w_{soft},w_{hard} ,s_i)
\end{equation}
where $w_{i+1}$ denotes the next to be generated word and $s_i$ denotes all the existing words generated in previous steps.
$w_{soft}$ is the set of learnable tokens that can be updated by the guidance during the generation process.
$w_{hard}$ is the hard prompt, which is randomly sampled from a predefined prompt set, such as ``Video showing'', ``Video of''.

To utilize the retrieved sentences, we transfer them into the above auto-regression process with our proposed retrieval-based loss $\mathcal{L}_{retrieval}$.
It consists of two granularities: the sentence-granularity $\mathcal{L}_{retrieval}^{S}$ and the word-granularity $\mathcal{L}_{retrieval}^{W}$ .
Specifically, the $\mathcal{L}_{retrieval}^{S}$ is formulated as
\begin{equation}~\label{sentence-granularity}
    \mathcal{L}_{retrieval}^{S} = -\sum_{\mathrm k=1}^{\mathrm N}\mathrm p\left(\mathrm w_k\right)\log\left(\mathrm q\left(\mathrm w_k\right)\right)
\end{equation}
where 
$\mathrm q\left(\mathrm w_k\right)$ is the probability of the $k$-th word in the vocabulary, and $\mathrm{p}(\cdot)$ denotes the probability of being soft target. 
$\mathrm{p}\left(\mathrm w_k\right)$ is computed as
\begin{equation}
    \mathrm{p}\left(\mathrm w_k\right) = \frac{1}{K} \sum_{R\in\mathcal{R}}\operatorname{AnglE}\left(R,s_{k}\right)
\end{equation}
where $K$ denotes the number of retrieved sentences, {$s_k$ represents the combination of the generated words in previous time steps with the $k$-th word}, $\mathrm{p}\left(\mathrm w_k\right)$ is the average of $\operatorname{AnglE}$'s alignment scores between $s_{k}$ and all the retrieved sentence in $\mathcal{R}$.
This loss adjusts learnable tokens so as to prompt GPT-2 to generate texts better reflect the content in the retrieved description of the given video.
%
For efficiency, we only calculate the scores for the top 100 candidate words based on the original probability distribution obtained by GPT-2.

On the other hand, considering that the retrieved sentences might contain noise,
%
we propose a word-granularity loss, which focuses on the high-frequency information in the retrieved sentences.
%
The word-granularity retrieval loss is formulated as 
\begin{equation}~\label{word-granularity}
    \mathcal{L}_{retrieval}^{W} = -\sum_{\mathrm k=1}^{\mathrm N}\mathrm {\hat{p}}\left(\mathrm w_k\right)\log\left(\mathrm q\left(\mathrm w_k\right)\right)
\end{equation}
\begin{equation}
    \mathrm{\hat{p}}\left(\mathrm w_k\right)= \frac{1}{L} \sum_{W\in\mathcal{W}}\operatorname{AnglE}\left(W,s_{k}\right)
\end{equation}
where $\mathcal{W}$ denotes all the high-frequency words in the retrieved sentences. $\mathrm{\hat{ p}}\left(\mathrm w_k\right)$ is the average of $\operatorname{AnglE}$'s alignment scores between $s_{k}$ and all the sampled high-frequency words in $\mathcal{W}$.

Inspired by~\cite{ZeroCap, ept, zerota}, we also employ two reliable losses to optimize the augment caption generation module.
The first loss aims to reduce the gap between the generated caption and the corresponding video:
\begin{equation} \label{clip_loss}
    \mathcal{L}_{vision} = -\sum_{\mathrm k=1}^{\mathrm N}(\frac{1}{T} \sum_{F\in\mathcal{F}}\operatorname{CLIP}\left(F,s_{k}\right))\log\left(\mathrm q\left(\mathrm w_k\right)\right)
\end{equation}
where $T$ is the total number of sampled video frames.
CLIP calculates the average matching score between each sampled frame $F$ in $\mathcal{F}$ and $s_k$.

The second loss maintains the natural language fluency of the generated caption while aligning it with the
video content:
\begin{equation}
    \mathcal{L}_{\text{language}}=\text{CE}\left(\mathrm{q}(w_{i+1}), \mathrm{q}(\hat{w}_{i+1})\right)
\end{equation}
\begin{equation}
    \hat{w}_{i+1} = \operatorname{PLM}(w_{hard} ,s_i)
\end{equation}
where CE is the cross-entropy loss function, $\hat{w}_{i+1}$ is the distribution of generated token without soft prompt.

\subsection{Inference} \label{sec: inference}
In our zero-shot setting, we directly perform test-time learning, where only text corpus of training split are used for retrieving caption candidates. The retrieved captions are used to generated soft targets as mentioned in Sec.~\ref{sec:captiongenerator}.
The learnable tokens are optimized by the total loss:
\begin{equation}
    \mathcal{L}_{total} = \lambda_l \mathcal{L}_{language} + \lambda_s \mathcal{L}_{retrieval}^{S} + \lambda_w \mathcal{L}_{retrieval}^{W} + \lambda_v \mathcal{L}_{vision}
\end{equation}
where $\lambda_l$, $\lambda_s$, $\lambda_w$ and $\lambda_v$ are the corresponding loss weights. 

Optimization occurs throughout the sentence generation, enhancing the correspondence between the generated sentence and video through the iterative application.
More specifically, the optimization is conducted at each step in the auto-regression process.  
When generating the stop token or reaching the maximum sentence generation length, we stop the process and initiate the generation of a new sentence while retaining the optimized soft prompt. 

\begin{figure}[!t]
\centering
\includegraphics[width=0.7\linewidth]{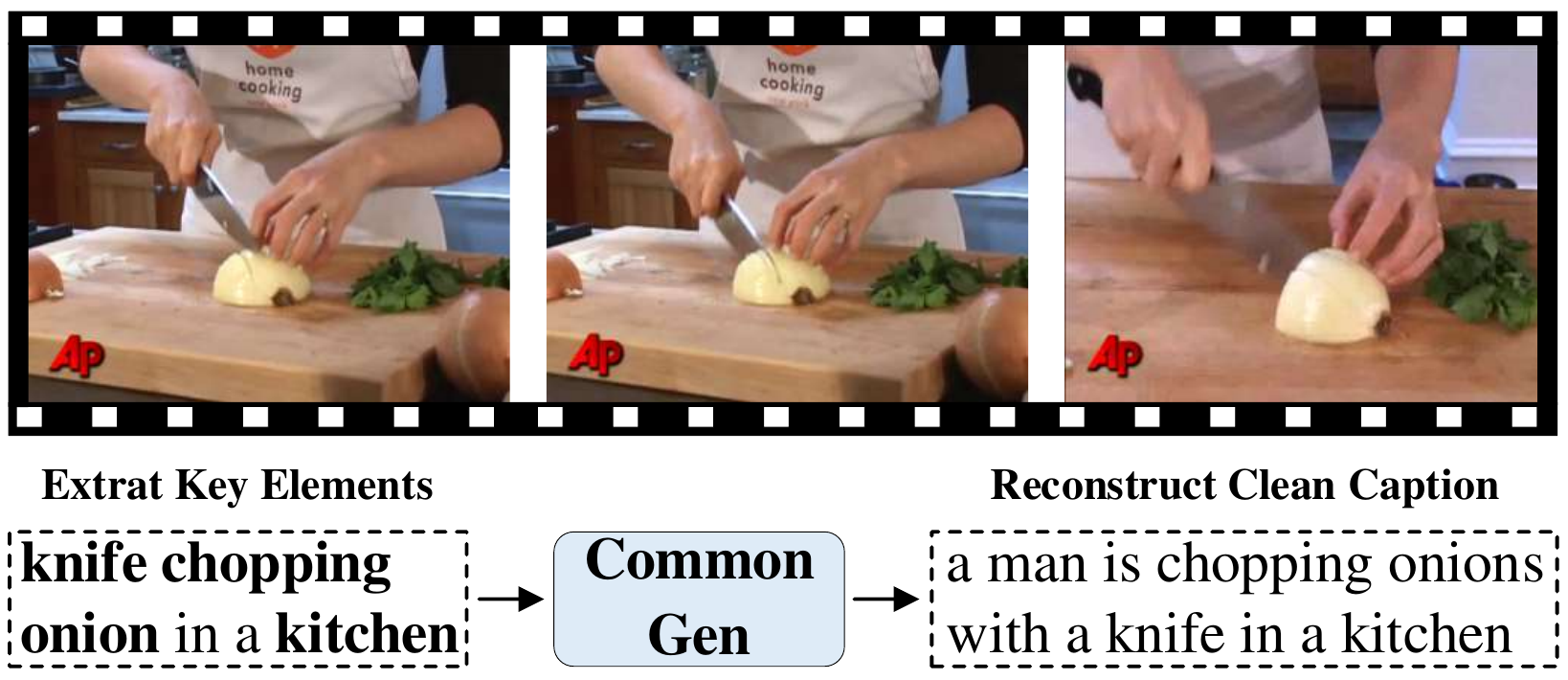}
\caption{Our sentence-cleaning post-process strategy {extract key elements, and use them to reconstruct the clean caption with the pre-trained CommonGen.}
}
\label{fig:post_process}
\end{figure}

{After all $M$ iterations are complete, we obtain $M$ generated captions. We calculate the similarity between these $M$ generated captions and the video as follows:
{
\begin{equation}
    s^* = \arg\max_{s_m \in \{s_1, s_2, \ldots, s_M\}} \left(\frac{1}{T} \sum_{F \in \mathcal{F}} \operatorname{CLIP}(F, s_m)\right).
\end{equation}
}%
The caption $s^*$ with the highest similarity is selected as the reliable caption.}
However, due to the limitations of zero-shot learning, the selected caption still contains some noise and lacks fluency.
We design a simple but effective sentence-cleaning strategy, which consists of two steps: (1) extracting essential elements like verbs and nouns from the selected sentence, and (2) employing a pre-trained keyword-sentence model CommonGen~\cite{CommonGen} to reconstruct these elements into a well-organized sentence.
CommonGen is a state-of-the-art model for constrained text generation, which takes keywords (such as ``knife'', ``chopping'', ``onion'' and ``kitchen'') as input and generates a more fluent sentence ``a man is chopping onions with a knife in a kitchen'', as shown in Figure~\ref{fig:post_process}. 
In the end, the cleaned sentence is used as the final result.

\section{Experiments}

\subsection{Datasets and Metrics}

\noindent\textbf{MSR-VTT}~\cite{msrvtt} is a large-scale video captioning dataset collected from a commercial video website. It contains 10,000 video clips covering 20 categories. Each clip has 20 English annotations. In the standard split, it includes 6,513 training videos, 497 validation videos, and 2,990 test videos.

\vspace{1mm}
\noindent\textbf{MSVD}~\cite{msvd} is a widely-use video captioning dataset, which is collected from Youtube. It consists of 1,970 video clips. Each clip has roughly 40 ground-truth captions. In the standard split, it includes 1,200 training videos, 100 validation videos, and 670 test videos.

\vspace{1mm}
\noindent\textbf{VATEX}~\cite{wang2019vatex} is a large-scale dataset that contains 34,991 videos with 10 English annotations. The standard split includes 25,910 training videos, 3,000 validation videos, and 6,000 test videos.

\vspace{1mm}
\noindent\textbf{Metrics}. In our experiments, we adopt four commonly used metrics to evaluate caption quality, including \textcolor{black}{BLEU@4 (B@4), METEOR (M), ROUGE-L (R), and CIDEr (C)}. Among these metrics, CIDEr is the main metric.

\subsection{Implementation Details}
We use the pre-trained XCLIP\footnote{https://huggingface.co/microsoft/xclip-base-patch16-kinetics-600-16-frames} as our video retriever. We uniformly sample 16 frames per video for video retrieval.
The visual feature extraction and text-image similarity calculation are performed using the pre-trained CLIP ViT-L/14\footnote{ https://openaipublic.azureedge.net/clip/models/b8cca3fd41ae0c99ba7e8951adf17d267cdb84cd88be6f7c2e0eca1737a03836/ViT-L-14.pt} model.
\textcolor{black}{Our video frame sampling strategy involves two steps: fixed-rate sampling followed by CLIP-based redundancy filtering. First, we uniformly sample three frames per second from the entire video. To reduce redundancy, we then apply a CLIP-based keyframe selection strategy. Starting with the first frame as an anchor, we compute the dot product similarity between each subsequent frame and the anchor using CLIP image embeddings. If the similarity is smaller than a predefined threshold $\lambda_{CLIP}$ (set to 0.9), the frame is added to the keyframe set $\mathcal{F}$ and becomes the new anchor. This process is repeated for the remaining frames.}
We use the pre-trained GPT-2 medium\footnote{https://huggingface.co/openai-community/gpt2-medium} as our caption generator, and 
the pre-trained AngLE\footnote{https://huggingface.co/WhereIsAI/UAE-Large-V1} to calculate the similarity between the generated caption and retrieval sentences.
Last, we leverage the pre-trained CommonGen\footnote{https://huggingface.co/mrm8488/t5-base-finetuned-common\_gen} to clean the generated sentences into fluent ones.
The aforementioned pre-trained models are publicly and readily available.
\textcolor{black}{All the pre-trained models used in our study adopt their original default parameters without any manual modifications, making them directly integrable into our system.}

Empirically, $\lambda_l$, $\lambda_v$, $\lambda_s$ and $\lambda_w$ are set to 1.6, 1.0, 0.8, and 0.3 respectively.
In our method, we set the number of learnable tokens to 5.
We set the number of retrieval sentences to 15 and the frequency size to 5.
We adopt the AdamW optimizer with a learning rate of 1e-4 and a weight decay weight of 0.3 to optimize our model.
Throughout the experiments, we employ 16 generation iterations per video.
To prevent sentences from being overly long and repetitive, we limit the number of tokens generated per sentence to 15.
\textcolor{black}{We implement our framework with PyTorch 1.13.1 and Transformers 4.32.1. All experiments are conducted on an Intel Xeon Platinum 8375C CPU, 64 GB of RAM, and one NVIDIA RTX A6000 GPU.}


\noindent\textbf{Discussion of Model Choice.}
\textcolor{black}{We follow many existing zero-shot captioning works~\cite{MERCap,MeaCap,DeCap,ZeroCap,MultiCapCLIP} and adopt CLIP as the visual feature extractor and GPT-2 as the text generator to remain consistent with prior approaches.
The CLIP model is pre-trained on 400 million image-text pairs, which plays the vision-language alignment role to map the image and sentence into a unified semantic space. 
The out-of-box GPT-2 is pre-trained on the large dataset WebText collected from a wide range of text data online. It is composed of several Transformer layers, which can cache the previous key and value tokens to infer the next token autoregressively. Therefore, GPT-2 is capable of generating text based on preceding content.
XCLIP is pre-trained on the Kinetics-600 dataset and has the zero-shot ability for video classification and video-text retrieval tasks. 
AnglE is a state-of-the-art model for calculating semantic textual similarity, which is able to be aware of differences among texts.
%
These pre-trained models offer strong zero-shot performance, are lightweight (parameter range: 194.93M–427.62M), easy to deploy, and well-suited for integration into a unified system for collaborative use. These factors make them a reasonable choice.
In our study, X-CLIP is used to retrieve reference sentences based on video content.  CLIP is responsible for visual feature extraction, frame sampling, and image-text matching to filter candidate tokens. GPT-2 handles text generation, and AnglE computes the similarity between candidate and reference sentences for token selection.}

\begin{figure}[h]
\centering
\includegraphics[width=0.7\linewidth]{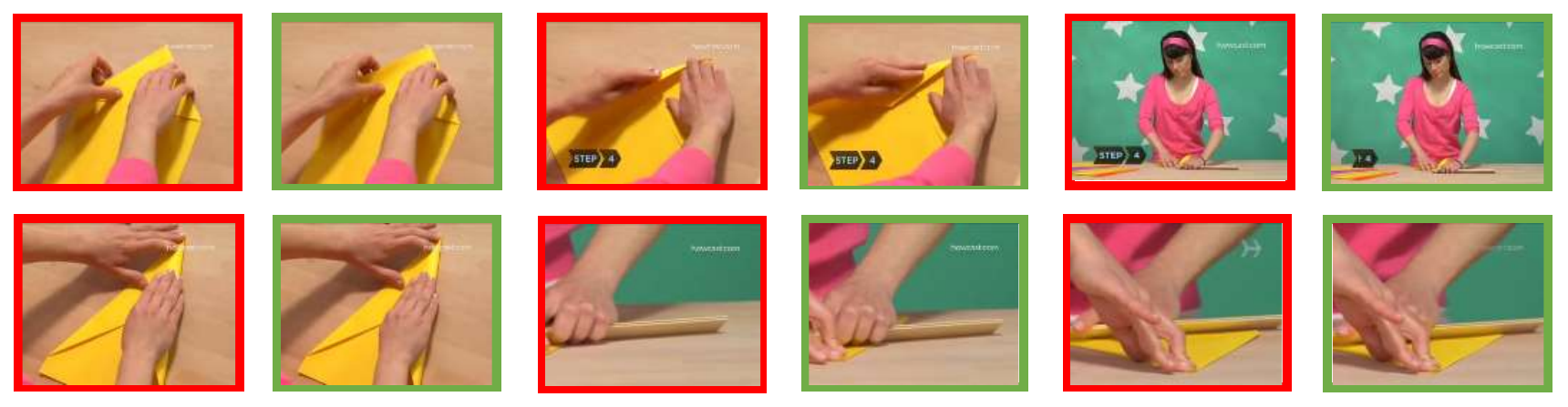}
\caption{Illustration of our CLIP-based sampling strategy. Selected frames are highlighted in red, while filtered (redundant) frames are marked in green. 
}
\label{fig:CLIP_sampling}
\end{figure}

\noindent\textbf{Discussion of Parameter Settings.}
\textcolor{black}{In this section, we provide a discussion on several important design choices. 
}
\textcolor{black}{For video frame sampling, a fixed sampling rate of 3 FPS is adopted, which balances efficiency and completeness for most videos (typically 15–30 FPS). For high-frame-rate videos (\textit{e.g.}, 60 FPS), a higher rate (\textit{e.g.}, 5 FPS) can optionally be used.
As illustrated in Figure~\ref{fig:CLIP_sampling}, setting the CLIP similarity threshold to $\lambda_{CLIP} = 0.9$ helps eliminate redundant frames while preserving critical visual content, making it a suitable default choice.}
\textcolor{black}{The discussion of other hyperparameter choices, such as the number of retrieved sentences, the number of learnable tokens, and the frequency size for the word-granularity loss, will be detailed in the Ablation Study section.}

\begin{table}[t!]
\centering
\caption[caption]{Results on MSVD across various models and supervision levels. \textcolor{black}{B@4, M, R, and C denote BLEU@4, METEOR, ROUGE-L, and CIDEr, respectively. $\uparrow$ indicates that higher is better. This notation applies to all subsequent tables.}}
\setlength{\abovecaptionskip}{-0.1cm}
\setlength{\belowcaptionskip}{-0.1cm}

\resizebox{0.7\textwidth}{!}{
\small
\begin{tabular}{ c|c|cccc}

\toprule
\multirow{2}{*}{Methods}&\multirow{2}{*}{Setting }&\multicolumn{4}{c}{Metrics}\\

 & &B@4 \textcolor{black}{$\uparrow$} &M \textcolor{black}{$\uparrow$}&R \textcolor{black}{$\uparrow$}&C \textcolor{black}{$\uparrow$} \\ 
\toprule
\multicolumn{6}{c}{Results on MSVD test set}\\
\toprule
VNS-GRU \cite{VNS_GRU}&\multirow{3}{*}{Supervised}&66.5&42.1&79.7&121.5\\
SemSynAN \cite{SemSynAN}&&64.4&41.9&79.5&111.5\\
HMN \cite{hmn}&&59.2&37.7&75.1&104.0\\
\toprule
ZeroCap~\cite{ZeroCap} &\multirow{5}{*}{Zero-shot}& 2.9 & 16.3 & 35.4 &9.6\\
MAGIC~\cite{MAGIC}  && {6.6} & 16.1 & 40.1  &14.0\\
EPT\cite{ept}     && 3.0 & 17.8 & 31.4 &17.4\\
Video-LLaMA\cite{videollama}     && 4.0 & 15.1 & 21.6 &1.3\\
RETTA(Ours) &&\textbf{23.3}&\textbf{28.5}&\textbf{56.4}&\textbf{49.8}\\
\toprule
\end{tabular}
}

\label{msvd_sota}
\vskip -0.1in
\end{table}

\begin{table}[t!]
\centering
\caption[caption]{Results on  MSRVTT across various models and supervision levels.}
\setlength{\abovecaptionskip}{-0.1cm}
\setlength{\belowcaptionskip}{-0.1cm}

\resizebox{0.7\textwidth}{!}{
\small
\begin{tabular}{ c|c|cccc}

\toprule
\multirow{2}{*}{Methods}&\multirow{2}{*}{Setting }&\multicolumn{4}{c}{Metrics}\\

 & &B@4 \textcolor{black}{$\uparrow$}&M \textcolor{black}{$\uparrow$}&R \textcolor{black}{$\uparrow$}&C \textcolor{black}{$\uparrow$} \\ 
\toprule
\multicolumn{6}{c}{Results on MSR-VTT test set}\\
\toprule

VNS-GRU \cite{VNS_GRU}&\multirow{3}{*}{
{Supervised}}&45.3&29.9&63.4&53.0\\
SemSynAN \cite{SemSynAN}&&46.4&30.4&64.7&51.9\\
HMN \cite{hmn}&&43.5&29.0&62.7&51.5\\
\toprule
ZeroCap~\cite{ZeroCap} &\multirow{10}{*}{
{Zero-shot}}& 2.3 & 12.9 & 30.4 &5.8\\
MAGIC~\cite{MAGIC}  && 5.5 & 13.3 & 35.4  &7.4\\
EPT\cite{ept}     && 3.0 & 14.6 & 27.7 &11.3\\
Video-LLaMA\cite{videollama}     && 4.9 & 16.8 & 25.3 &2.3\\
ZS-CapCLIP\cite{clip}     && 4.0 & 15.0 & 31.0 &5.0\\
MultiCapCLIP\cite{MultiCapCLIP}     && 13.3 & 19.5 & \textbf{43.3} & 15.5\\
{DeCap-BookCorpus~\cite{DeCap}}&&{6.0}&{12.7}&{-}&12.3\\
DeCap-CC3M~\cite{DeCap} &&6.2&14.9&-&{15.0}\\
DeCap-COCO~\cite{DeCap} &&\textbf{14.7}&\textbf{20.4}&\textbf{-}&18.6\\
RETTA(Ours) &&{14.0}&{19.3}&{42.2}&\textbf{24.3}\\
\toprule
\end{tabular}
}

\label{msrvtt_sota}
\vskip -0.1in
\end{table}%

\begin{table}[t!]
\centering
\caption[caption]{Results on VATEX across various models and supervision levels.}
\setlength{\abovecaptionskip}{-0.1cm}
\setlength{\belowcaptionskip}{-0.1cm}

\resizebox{0.7\textwidth}{!}{
\small
\begin{tabular}{ c|c|cccc}

\toprule
\multirow{2}{*}{Methods}&\multirow{2}{*}{Setting }&\multicolumn{4}{c}{Metrics}\\

 & &B@4 \textcolor{black}{$\uparrow$}&M \textcolor{black}{$\uparrow$}&R \textcolor{black}{$\uparrow$}&C \textcolor{black}{$\uparrow$} \\ 
\toprule
\multicolumn{6}{c}{{Results on VATEX public test set}}\\
\toprule
{VaTeX \citep{wang2019vatex}}&\multirow{2}{*}{Supervised}&28.4&21.7&-&45.1\\
HMN \cite{hmn}&&33.2&22.7&49.3&51.6\\
\toprule
{DeCap-BookCorpus~\cite{DeCap}}&\multirow{5}{*}{Zero-shot}&{4.1}&{9.9}&{-}&11.8\\
{DeCap-CC3M~\cite{DeCap}}&&7.3&12.6&-&18.4\\
{DeCap-COCO~\cite{DeCap}}&&\textbf{13.1}&15.3&\textbf{-}&{18.7}\\
Video-LLaMA\cite{videollama}     && 4.3 & {16.3} & 21.8 &3.8\\
RETTA(Ours) &&{11.4}&\textbf{16.3}&\textbf{32.6}&\textbf{23.8}\\
\toprule
\end{tabular}
}

\label{vatex_sota}
\vskip -0.1in
\end{table}

\subsection{Performance Comparison}
We compare our method with several state-of-the-art models under different settings on three public benchmark datasets.
The results are shown in Table~\ref{msvd_sota}, \ref{msrvtt_sota} and \ref{vatex_sota}.
We note that the performance of supervised algorithms is significantly higher than zero-shot methods, but they require expensive annotation costs.
Compared to the other zero-shot methods, our method achieves the best performance in terms of the main metric CIDEr on all the three datasets.
%
As shown in Table~\ref{msvd_sota}, our method achieves a CIDEr score of 49.8 on the MSVD benchmark, leading by about an absolute 32.4 over the previous best result.
On the more challenging benchmark MSR-VTT, our method also boosts the CIDEr score by about 30.6\%. 
%
On the VATEX dataset, our method achieves a 23.8 CIDEr Score, gaining 27.3\%  improvement in terms of the main metric CIDEr on VATEX.
\textcolor{black}{
While our method underperforms slightly on traditional textual metrics such as BLEU@4, METEOR, and ROUGE-L compared to DeCap-COCO and MultiCapCLIP on MSRVTT and VATEX. 
This is primarily because the latter methods are trained on the full set of reference captions from the MS COCO dataset, thus benefiting from more extensive language supervision.
}
\textcolor{black}{
These metrics are better suited for machine translation and may not fully reflect the relevance of captions to video content.
In contrast, our task focuses on describing video content, and the notable improvement in CIDEr—a metric specifically designed to measure vision-text relevance—demonstrates the effectiveness of our method in video captioning.
}
\textcolor{black}{Overall, our method performs best on MSVD (CIDEr 49.8), followed by MSRVTT (24.3) and VATEX (23.8).
The performance differences mainly stem from the varying complexity of the datasets: MSVD is relatively simple with 1,970 video clips; MSRVTT includes around 10,000 videos with greater diversity; and VATEX, the most challenging, contains 34,991 videos covering 600 action categories. Our method achieves the strongest CIDEr scores across all three datasets, with particularly notable performance on MSVD (49.8). While MSRVTT and VATEX present greater diversity and complexity, our approach still achieves the best CIDEr scores (24.3 and 23.8, respectively), showing robust generalization.
This consistent improvement on CIDEr demonstrates that our method effectively captures video content and generates high-quality captions even under diverse scenarios.}

\textcolor{black}{In Table~\ref{tab:compute_comparison}, we also provide a quantitative comparison of training compute (estimated by GPU memory consumption during training), parameter size, and inference time between our proposed RETTA and the other zero-shot video captioning methods. 
Both EPT and our RETTA method are test-time adaptation approaches, unlike DeCap-CC3M, which involves extensive pretraining on textual data ((\textit{e.g.}, 25,197 seconds on CC3M)) but offers very fast inference (0.1s per sample). 
RETTA skips pretraining and performs adaptation during testing. Despite a longer inference time, RETTA delivers a better CIDEr score (23.8 \textit{v.s.} 18.4).
\textcolor{black}{
Thus, DeCap-CC3M is more suitable for scenarios where low inference latency is critical (\textit{e.g.}, real-time applications) and higher training cost is acceptable.}
Furthermore, although RETTA introduces additional overhead due to the integration of four pre-trained models (with an absolute increase of 5334 MB in training memory, 530.07M in parameter size, and 17.8s in inference time compared to EPT), it achieves a significant improvement in CIDEr score (23.8 \textit{v.s.} 8.9), demonstrating a favorable trade-off between cost and performance.}
\textcolor{black}{
Therefore, for latency-insensitive scenarios (\textit{e.g.}, offline batch processing), both RETTA and EPT are viable choices. In particular, RETTA is better suited when higher performance is required and a text retrieval corpus is available, such as offline summarization of vertical-domain videos.}

\begin{table}[t!]
  \centering
  \caption{Efficiency and performance comparison. MB, M, and s represent megabytes of GPU memory usage, millions of model parameters, and seconds of inference time, respectively.}
  \label{tab:compute_comparison}
  \resizebox{0.8\textwidth}{!}{ 
  \begin{tabular}{lccccc}
    \toprule
    Method & GPU Memory Usage
 (MB) $\downarrow$ & Parameters Size (M) $\downarrow$ & Training Time (s) $\downarrow$ & Inference Time (s) $\downarrow$ & CIDEr $\uparrow$ \\
    \midrule
    DeCap-CC3M & 9656 & 219.41 & 25197 & 0.1 & 18.4\\
    EPT & 4408 & 782.44 & 0 & 42.6 & 8.9 \\
    RETTA(Ours) & 9742 & 1312.51 & 0 & 60.4 & 23.8 \\
    \bottomrule
  \end{tabular}
  }
\end{table}

\subsection{Ablation Study}
In this section, we perform a set of ablation studies on VATEX as it is the largest and most challenging.

\begin{table}[!tbp]
	\centering
  \caption{Performance with different our proposed component. Experiments are conducted on the VATEX dataset. \textcolor{black}{ $\mathcal{L}_{retrieval}^{S}$, $\mathcal{L}_{retrieval}^{W}$, and  $\mathcal{S}_{clean}$ represent our proposed losses at the sentence and word levels, as well as the sentence-cleaning strategy, respectively.}}
	\label{granularity}
      \resizebox{ 0.7\linewidth}{!}
{\setlength{\tabcolsep}{1mm}{
    \begin{tabular}{lccccccc}
	\toprule
	 \#  &$\mathcal{L}_{retrieval}^{S}$&$\mathcal{L}_{retrieval}^{W}$ & $\mathcal{S}_{clean}$ & B@4 & M & R & C\\ 
	\midrule
	  1& &  & & 1.4 & 9.8 & 19.4 & 8.9\\
  2& \checkmark & & & 2.5  & {13.3}  & 22.6 & {20.8}\\
	 3& & \checkmark & & 2.6  & {12.2}  & 22.1 & {14.9}\\
	 4& \checkmark& \checkmark & & {2.4} & 13.5 & 22.9 & 21.4\\
         5&\checkmark & \checkmark&\checkmark & 11.4 & 16.3 & 32.6 & 23.8\\
	  \bottomrule
	\end{tabular}}}
 \vspace{-5mm}
\end{table}%

\noindent\textbf{Effect of our proposed component.}
To fully analyze the effect of each proposed component, we conduct experiments on the challenge VATEX.
As shown in Table~\ref{granularity}, we observe that each component assists the baseline to get better performance on all metrics.
For example, Row~2 shows that using only sentence-granularity retrieval loss leads to a CIDEr score increase of $133.7\%$ compared to the baseline in Row~1. In contrast, employing word-granularity retrieval loss results in a CIDEr enhancement of $67.4\%$ and a BELU@4 score improvement of $85.7\%$ (Row~3 \vs ~Row~1). The second-best performance is achieved when both types of losses are used together (Row~4 \vs ~Row~1).
In the comparison, the results with $\mathcal{S}_{clean}$ post-processing reach a further improvement of $11.2\%$ in the term of CIDEr (Row5 \vs~ Row4).

\vspace{1mm}


\noindent\textbf{Effect of different retrieval corpora.} 
In Table~\ref{Retrieval Corpus}, we compare the performance of our method under different retrieval corpora.
On the MSR-VTT benchmark, the CIDEr score increases around $9.9\%$ by leveraging the TestSet as the retrieval corpus instead of the TrainSet (Row 1 \vs~Row 2).
On MSVD and VATEX datasets, employing TestSet improves around $21.9\%$ (Row 3 \vs~Row 4) and $8.4\%$ (Row 5 \vs~Row 6) compared to TrainSet.
This demonstrates that different knowledge corpora can significantly impact performance, especially on relatively small datasets (\textit{e.g.}, MSVD).
%
In this paper, unless otherwise specified, we use TrainSet as our default text corpus.
\vspace{1mm}

\noindent\textbf{Effect of video retriever.}
In order to study the impact of video retrievers on our method, we use different configurations of XCLIP as our retriever.
The results are present in Table~\ref{retriever}.
We observe that when XCLIP utilizes a ViT with a size of B/16 as the backbone, it achieves an approximate improvement of 2.2\% in the CIDEr score compared to using B/32 (Row 2 \vs~Row 1).
The Kinetics-600 dataset is the extended version of Kinetics-400, containing more training data.
Therefore, XCLIP pre-trained on Kinetics-600 brings a further improvement of 1.7\% compared to XCLIP pre-trained on Kinetics-400 (Row 3 \vs~Row 2).
Row 4 achieves the best performance on all metrics. It is trained with 16 frames per video and outperforms Row 3 using 8 frames.
Therefore, in the other experiments conducted in this paper, we use the XCLIP in Row 4 as our default retriever.

\begin{figure*}[!ht]
\centering
\resizebox{0.95\textwidth}{!}{ 
\begin{minipage}{\textwidth}

\begin{subfigure}[b]{0.49\linewidth}
    \includegraphics[width=\linewidth]{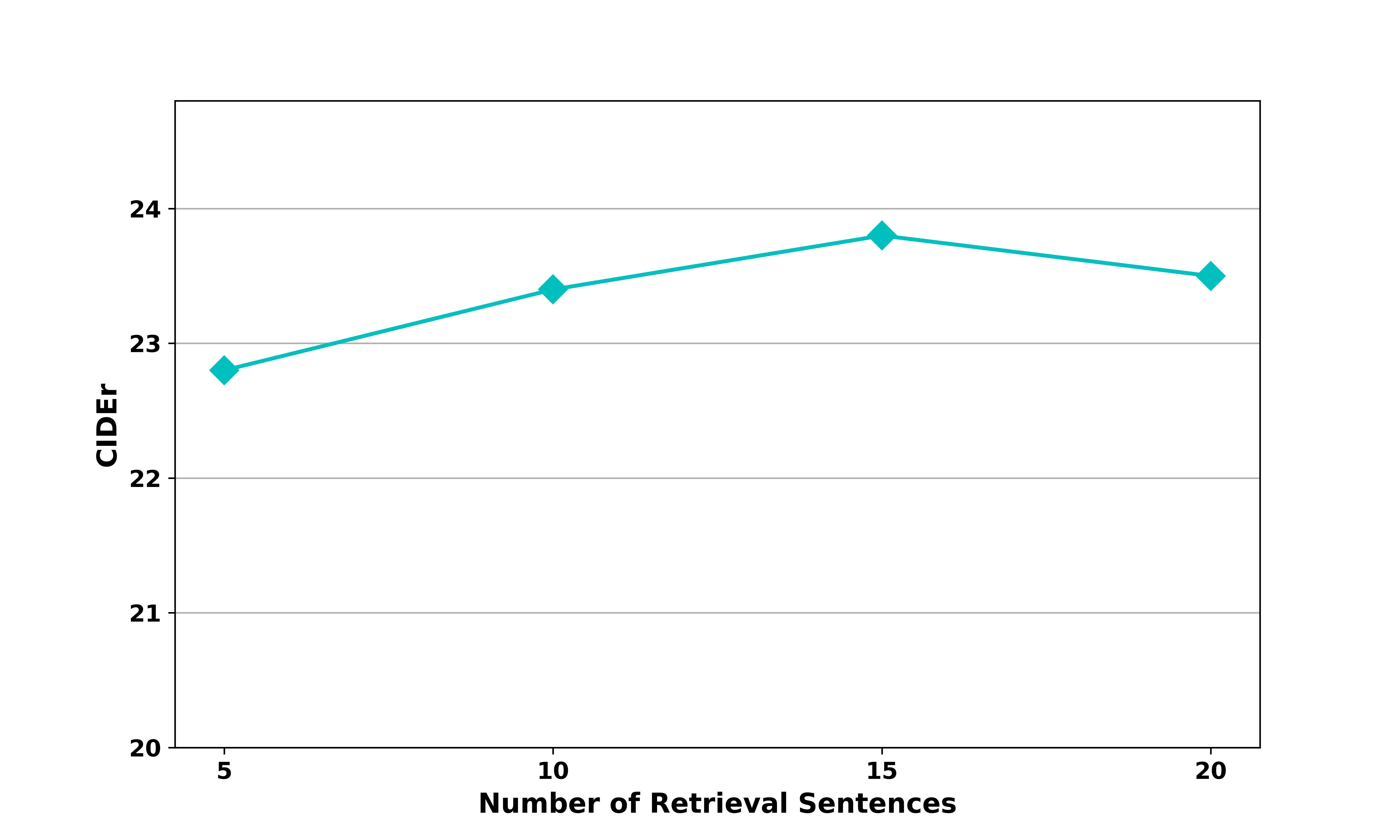}
    \caption{}
    \label{cider_retrieved_sentences}
\end{subfigure}
\hfill
\begin{subfigure}[b]{0.49\linewidth}
    \includegraphics[width=\linewidth]{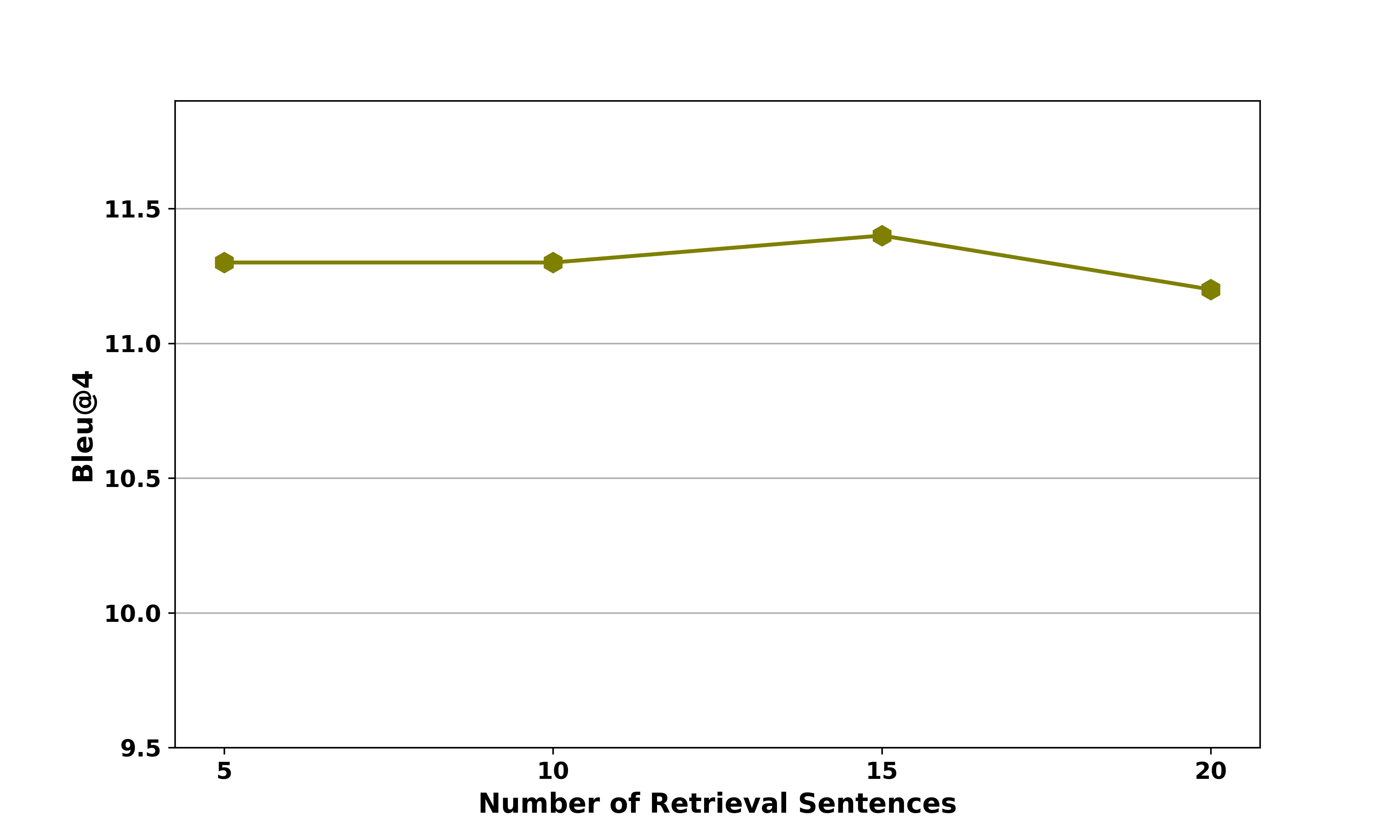}
    \caption{}
    \label{bleu_retrieved_sentences}
\end{subfigure}

\vspace{0.1mm}

\begin{subfigure}[b]{0.49\linewidth}
    \includegraphics[width=\linewidth]{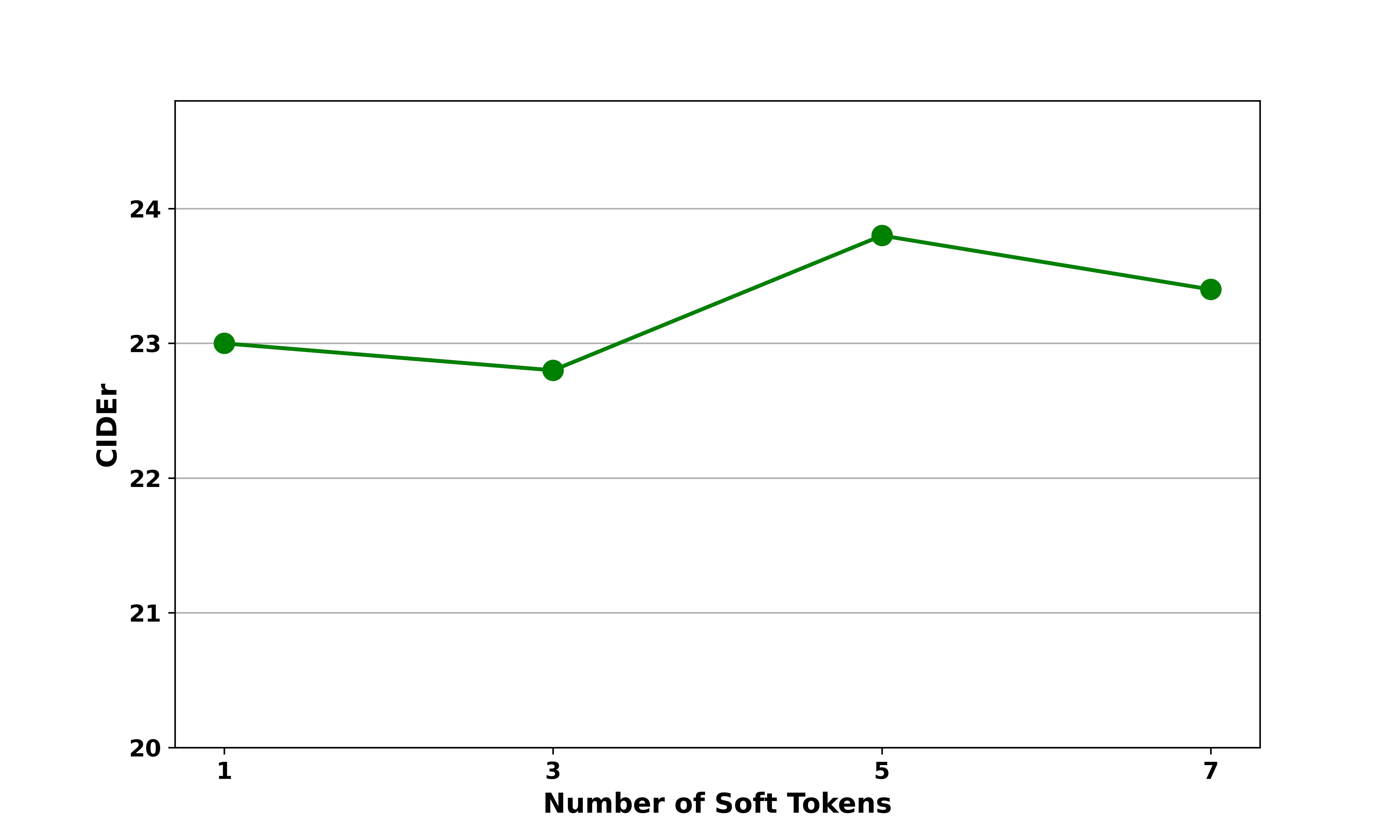}
    \caption{}
    \label{cider_soft_tokens}
\end{subfigure}
\hfill
\begin{subfigure}[b]{0.49\linewidth}
    \includegraphics[width=\linewidth]{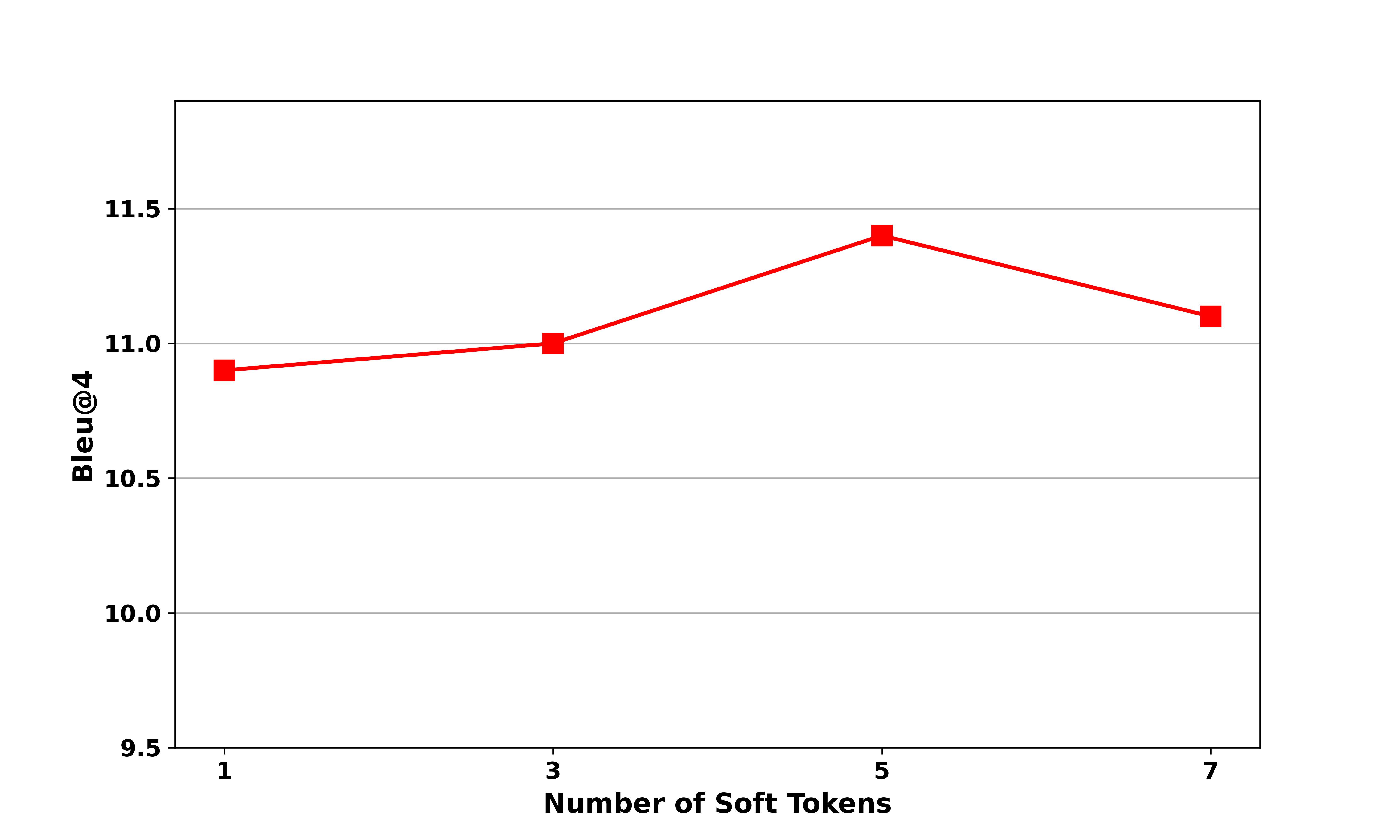}
    \caption{}
    \label{bleu_soft_tokens}
\end{subfigure}

\vspace{0.1mm}

\begin{subfigure}[b]{0.49\linewidth}
    \includegraphics[width=\linewidth]{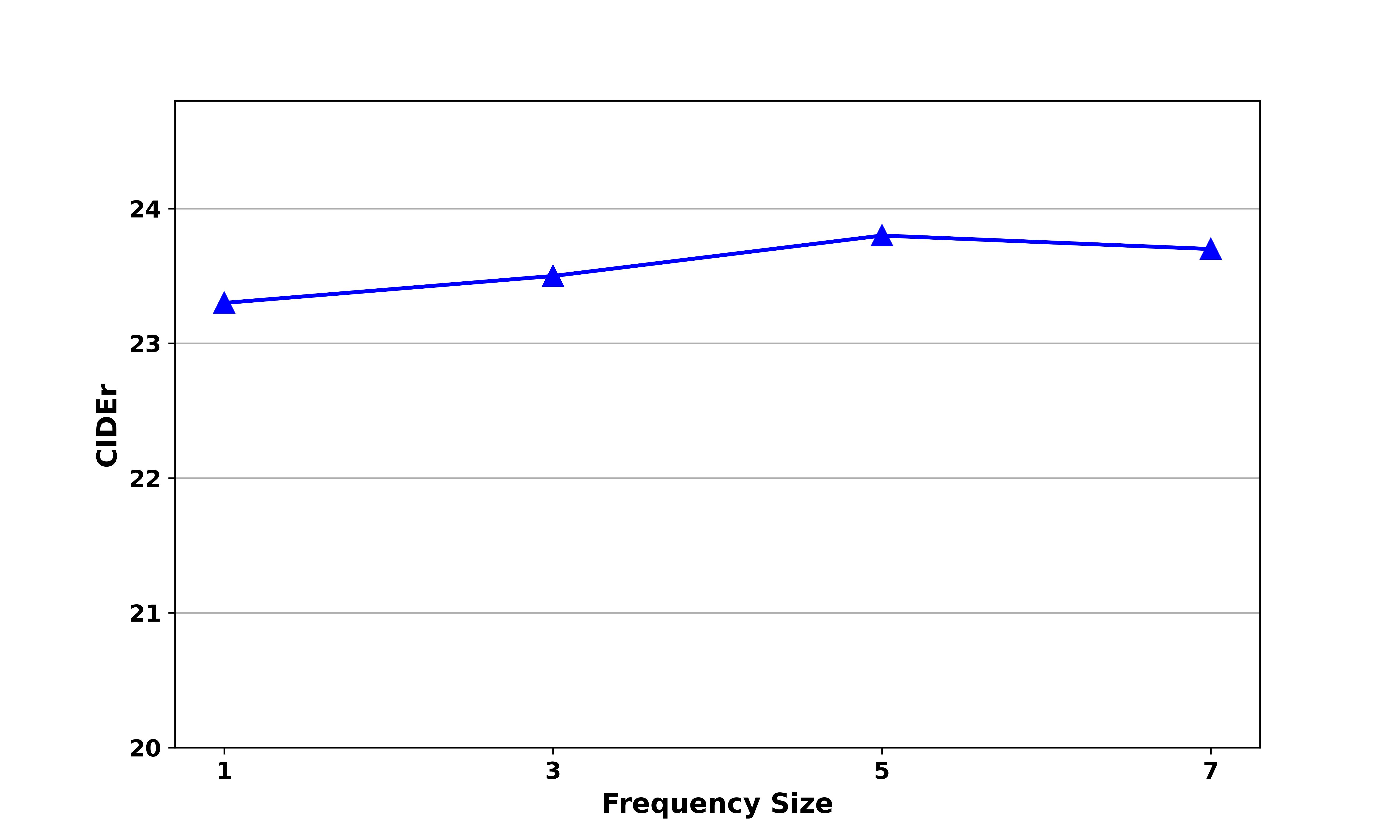}
    \caption{}
    \label{cider_frequency_size}
\end{subfigure}
\hfill
\begin{subfigure}[b]{0.49\linewidth}
    \includegraphics[width=\linewidth]{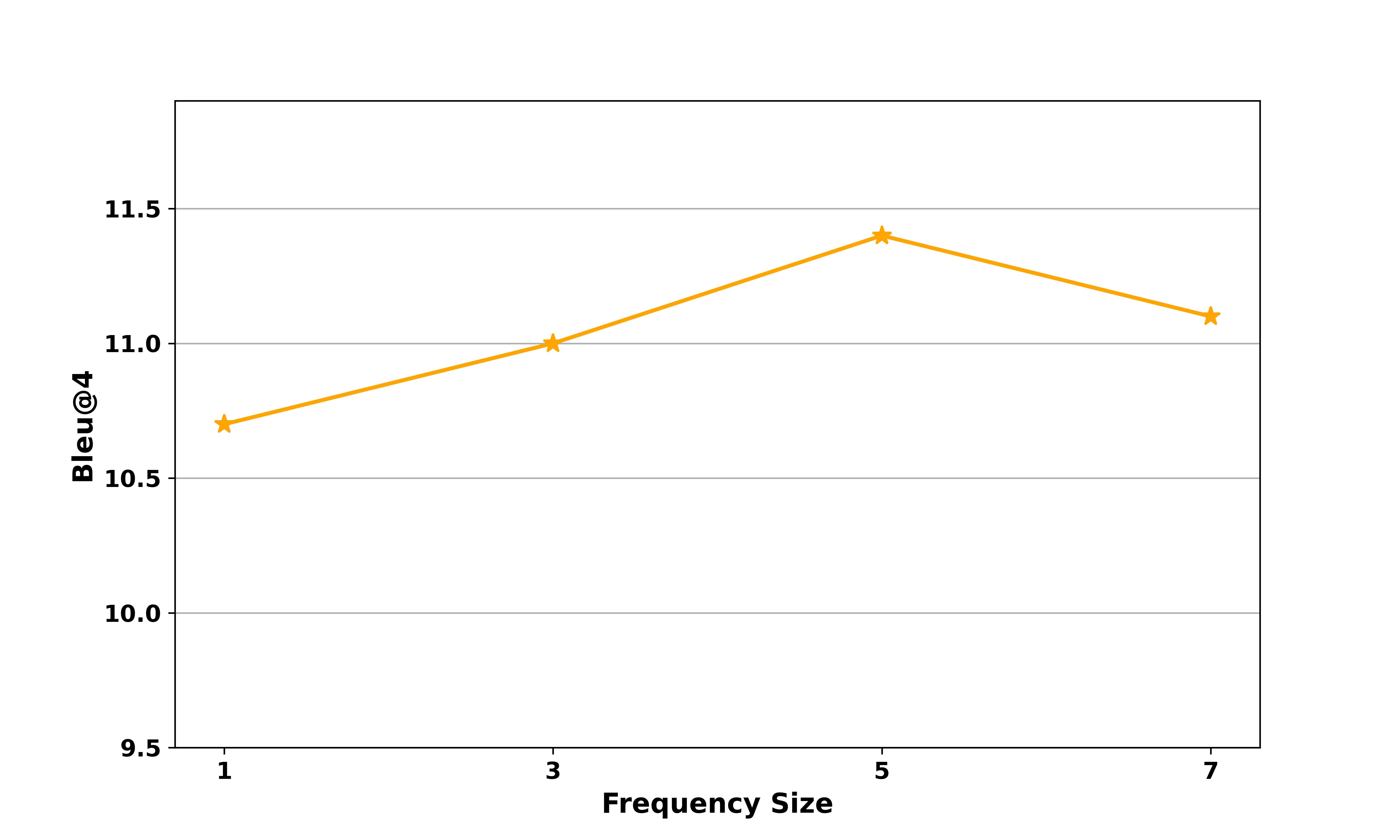}
    \caption{}
    \label{bleu_frequency_size}
\end{subfigure}

\vspace{0.1mm}

\begin{subfigure}[b]{0.49\linewidth}
    \includegraphics[width=\linewidth]{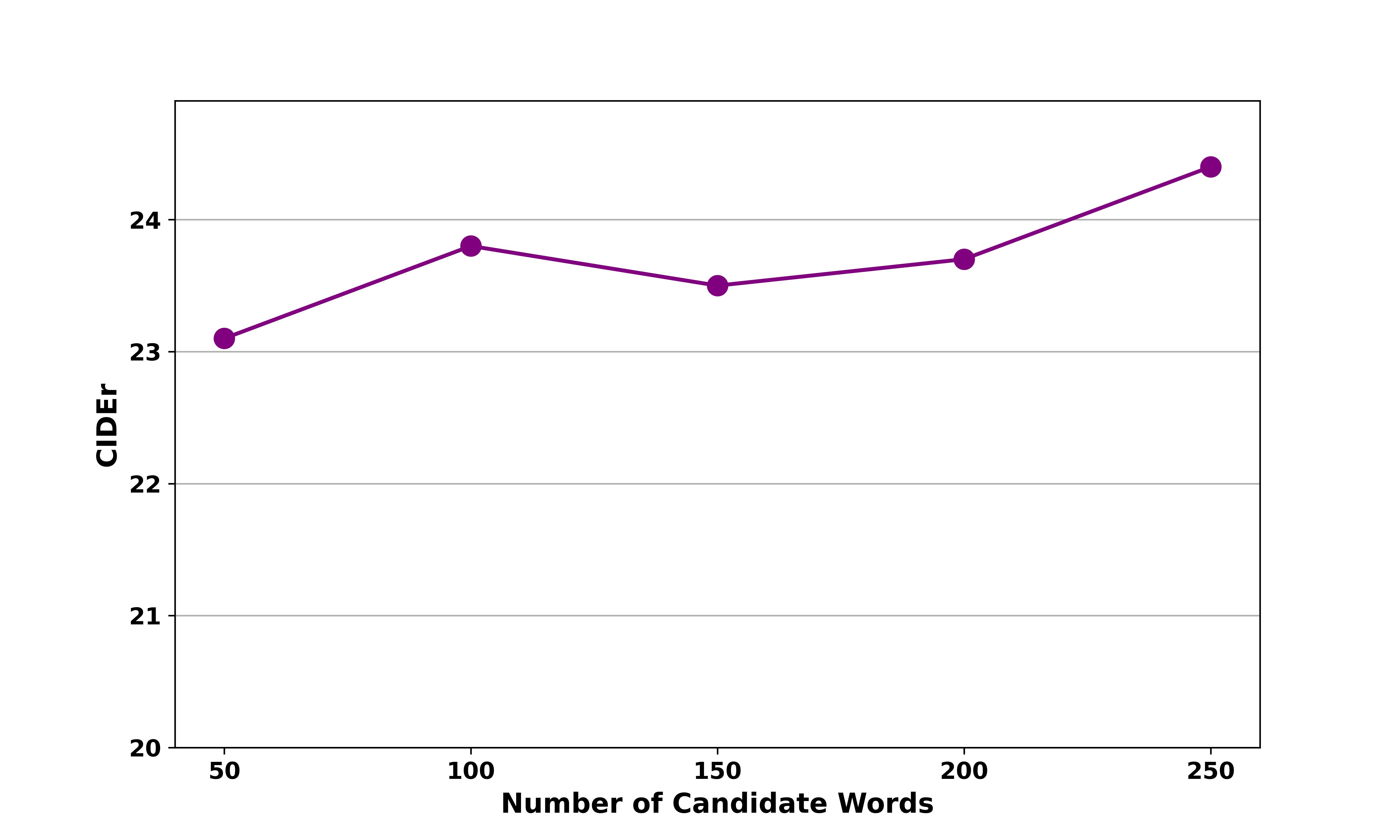}
    \caption{}
    \label{fig:cider_candidate_tokens}
\end{subfigure}
\hfill
\begin{subfigure}[b]{0.49\linewidth}
    \includegraphics[width=\linewidth]{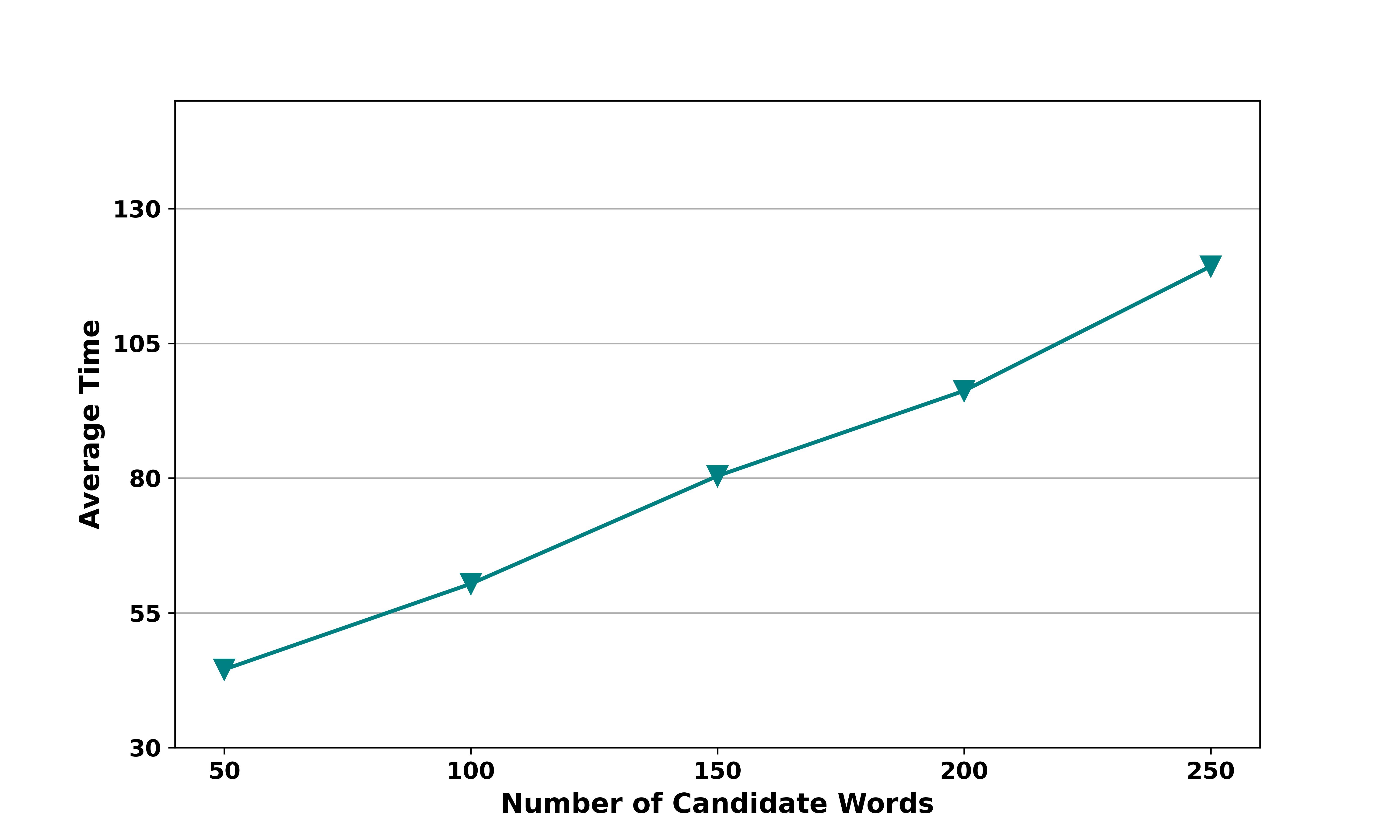}
    \caption{}
    \label{fig:time_candidate_tokens}
\end{subfigure}
\end{minipage}
} 

\caption{Performance comparison on the VATEX dataset under different settings: (a)–(b) retrieved sentence count, (c)–(d) soft token count, (e)–(f) frequency size, and (g)–(h) candidate word count.}
\label{fig:combined_vatex_analysis}
\end{figure*}

\vspace{1mm}
\noindent\textbf{Effect of the number of retrieved sentences.} 
We study the effect of using different numbers of retrieved sentences.
Figure~\ref{fig:combined_vatex_analysis} (a) and (b) show the difference in video captioning performance as $K$ varies from 5 to 20.
We find that retrieving 15 sentences offers better CIDEr scores and BLEU@4 scores compared to other numbers.
It can be concluded that too many or too few retrieval sentences can degrade the performance.
This is because too many retrieved sentences may introduce too much noise while too few retrieved sentences cannot provide sufficient information.
\textcolor{black}{A similar observation was made in~\cite{vtar}, which notes that additional retrieved sentences may bring in unrelated descriptions and semantic noise, ultimately harming caption quality. Therefore, setting K=15 results in the best captioning performance. }

\begin{table}[!tbp]
	\centering
  \caption{Effect of retrieval corpora. The experiments are conducted on test split of MSVD, MSR-VTT and VATEX datasets. }
	\label{Retrieval Corpus}
      \resizebox{ 0.7\linewidth}{!}
{\setlength{\tabcolsep}{1mm}{
    \begin{tabular}{clccccc}
	\toprule
	Dataset & \# &Retrieval Corpus & B@4 & M & R & C\\ 
	\midrule
	\multirow{2}{*}{MSR-VTT} & 1 &TrainSet & 14.0     & 19.3     & 42.2     & 24.3\\
 & 2 &TestSet & 15.3     & 20.2     & 43.1    & 26.7 \\
	\midrule
	\multirow{2}{*}{MSVD} & 3 &TrainSet & 23.3    & 28.5     & 56.4    & 49.8\\
 & 4 &TestSet  & 29.4     & 31.8    & 60.0     & 60.7\\
        \midrule
	\multirow{2}{*}{VATEX} & 5 &TrainSet  & 11.4     & 16.3     & 32.6     & 23.8 \\
 & 6 &TestSet & 12.3     & 17.1     & 33.2   & 25.8\\
	  \bottomrule
	\end{tabular}}}
\end{table}%


\begin{table}[!tbp]
	\centering
  \caption{Performance with different configuration XCLIP. The experiments are conducted on the VATEX dataset. }
	\label{retriever}
        \resizebox{ 0.65\linewidth}{!}
{\setlength{\tabcolsep}{1mm}{
    \begin{tabular}{lccccccc}
	\toprule
	 \# &Backone & $N$ & Pre-train & B@4 & M & R & C\\
	\midrule
	  1 &  VIT-B/32 &8 & Kinetics-400 & 10.5& 15.6 & 31.9& 22.3\\
  2 & VIT-B/16 & 8 & Kinetics-400 & 10.9  & {15.8}  & 32.2  & {22.8}\\
  3 & VIT-B/16 & 8 & Kinetics-600 & 11.2  & {16.1}  & 32.3  & {23.2}\\
	 4 & VIT-B/16 &16 & Kinetics-600 & {11.4} &16.3 & 32.6 & 23.8\\
	  \bottomrule
	\end{tabular}}}
\end{table}%

\noindent\textbf{Effect of the number of learnable tokens.}
We study the effect of the number of learnable tokens, and Figure~\ref{fig:combined_vatex_analysis} (c) and (d) show that 5 is the better choice than other numbers.
In Figure~\ref{fig:combined_vatex_analysis} (c), the CIDEr score initially decreases as the number increases from 1 to 3, then rises to its highest point from 3 to 5, followed by another decrease from 5 to 7.
In Figure~\ref{fig:combined_vatex_analysis} (d), the Bleu@4 score first increases as the number goes from 1 to 5, then decreases from 5 to 7.
Therefore, we use 5 as our default number of soft tokens.
\textcolor{black}{A similar choice was made in the work~\cite{tpt}, which uses 4 learnable tokens.}

\noindent\textbf{Effect of the frequency size of the word-granularity loss.}
In Figure~\ref{fig:combined_vatex_analysis} (e) and (f), we conduct experiments on VATEX to explore the impact of the frequency size $L$ of word-granularity loss.
From the Figure~\ref{fig:combined_vatex_analysis} (e), we observe that the CIDEr score first rises before reaching the moderate number (\ie, $L=5$), and then starts to decrease slightly.
The main reason is that when $L$ is large, some noisy words are introduced.
On the contrary, if $L$ is small, some essential words may be neglected.
In Figure~\ref{fig:combined_vatex_analysis} (f), a similar trend is observed in the BLEU metric as well.
\textcolor{black}{A similar choice was also made in~\cite{MERCap} when selecting the number of retrieved entities. Moreover, comparative results show that increasing the number of retrieved entities beyond 5 fails to improve metrics and instead causes performance to drop.}
Therefore, we select an appropriate number (\ie, $L=5$) as the frequency size. 


\vspace{1mm}

\noindent\textbf{Effect of the number of candidate words }
As mentioned in Sec.~\ref{sec:captiongenerator}, to enhance efficiency, we compute the loss using the top 100 candidate words based on the original probability distribution obtained by the language model.
In Figure~\ref{fig:combined_vatex_analysis} (g) and Figure~\ref{fig:combined_vatex_analysis} (h), we conduct experiments on VATEX to study the impact of the number of candidate words.
%
From Figure~\ref{fig:combined_vatex_analysis} (g), we find that using the top 250 candidate tokens achieves the highest CIDEr score compared to other numbers, with the top 100 candidate tokens achieving the second-best CIDEr score.
In contrast, in Figure~\ref{fig:combined_vatex_analysis} (h), we observe that the average time consumed per video increases roughly linearly with the number of candidate tokens.
Compared to using 100 tokens, using 250 tokens takes about twice as long.
To balance performance and time consumption, we set the number of candidate words to 100.


\noindent\textbf{Effect of hard prompt set }
In our method, we randomly sample a phrase from a predefined set as the hard prompt.
The set we designed includes three phrases: ``Video showing'', ``Video of'', and ``Video shows''.
In Table~\ref{hard_prompt}, we compare the performance of our method under different sizes of hard prompt sets. 
Row~1 displays the worst CIDEr score in the absence of any hard prompts.
Using ``Video showing'' as a hard prompt consistently results in a CIDEr improvement of about 1.7\% (Row~2 \vs~Row~1).
The inclusion of ``Video showing'' and ``Video of'' in the hard prompt set leads to a CIDEr increase of around 0.8\% (Row~3 \vs~Row~2).
With the addition of ``Video shows'', the CIDEr score in Row~4 leads by an absolute 0.3 over Row~3.
%
Therefore, in our other experiments, the phrases covered in Row~4 are set as our default hard prompt set.
\textcolor{black}{
Inspired by ZeroCap’s use of the manually crafted ``Image of'' prompt~\cite{ZeroCap}, our predefined set of hard prompts was also manually designed and consists of three simple and intuitive phrases.
\textcolor{black}{
During inference, following prior work~\cite{zerota}, the hard prompts are sampled using a uniformly random strategy.}
Table~\ref{tab:different prompt sampling strategies} compares the effect of different sampling strategies, including fixed sampling of each specific prompt and uniformly random sampling.
We find that using any of the three individual fixed prompts yielded very similar performance (CIDEr scores of 23.2–23.3), indicating that the model does not show a strong preference for any specific phrase. Thus, weighting among prompts is unnecessary. 
\textcolor{black}{
Moreover, random sampling achieves a mean of 24.0 and a standard deviation of 0.3, reflecting a stable and slight improvement over fixed prompts. Supported by prior studies~\cite{zerota} and our experimental results, this strategy appears to be a reasonable choice.
}
}

\begin{table}[h]
	\centering
  \caption{Performance with different sizes of hard prompt sets. Experiments are conducted on VATEX.}
	\label{hard_prompt}
      \resizebox{ 0.6\linewidth}{!}
{\setlength{\tabcolsep}{1mm}{
    \begin{tabular}{c  ccc  cccc}
   \toprule
	 
  \multirow{3}[3]{*}{\#} & \multicolumn{3}{c}{Hard Prompt} & \multicolumn{4}{c}{Metrics}      \\
           &\multicolumn{1}{c}{Video } &Video &Video      & \multirow{2}{*}{B@4} & \multirow{2}{*}{M}     & \multirow{2}{*}{R}     & \multirow{2}{*}{C}      \\
            &\multicolumn{1}{c}{showing} &of&shows      &  &      &      &       \\            
\midrule
	 1&&&&10.7&16.0&32.0&22.9 \\
  2&\checkmark&&&10.9&16.1&32.3&23.3 \\
  3&\checkmark&\checkmark&&11.2&16.4&32.8&23.5 \\
  4&\checkmark&\checkmark&\checkmark&11.4&16.3&32.6&23.8 \\

	  \bottomrule
	\end{tabular}}}
\end{table}%

\begin{table}[htbp]
  \centering
  \caption{Comparisons of CIDEr performance under different prompt sampling strategies. For random sampling, we report the mean and standard deviation across multiple runs.}
  \label{tab:different prompt sampling strategies}
  \resizebox{1.0\textwidth}{!}{ 
    \begin{tabular}{lccccccc}
      \toprule
      Strategy & Fix sampling (video showing) & Fix sampling (video of) & Fix sampling (video shows) & Uniformly random sampling \\
      \midrule
      CIDEr     & 23.3  & 23.2  & 23.2  & 24.0 $\pm$ 0.3   \\
      \bottomrule
    \end{tabular}
  }
\end{table}

\noindent\textbf{Zero-shot image captioning.}
Our method can be adapted for zero-shot image captioning by 
swapping XCLIP with CLIP.
In Table~\ref{mscoco_sota}, we compare the performance among several zero-shot image captions methods on the MS-COCO~\cite{MS-COCO} dataset.
We achieve the second-best results on all metrics, only falling behind MAGIC~\cite{MAGIC}.
This is because MAGIC fine-tuned the language model on the text corpus of MS-COCO captions.
Compared to ZeroCap~\cite{ZeroCap} and EPT~\cite{ept}, our CIDEr score improves 
 by an absolute 31.9 and 27.8, respectively.
\begin{table}[t!]
\centering
\caption[caption]{Image captioning results on MS-COCO. }

\resizebox{0.6\textwidth}{!}{
\small
\begin{tabular}{ c|c|cccc}

\toprule
\multirow{2}{*}{Methods}&\multirow{2}{*}{Setting }&\multicolumn{4}{c}{Metrics}\\

 & &B@4&M&R&C \\ 
\toprule
\multicolumn{6}{c}{Results on MS-COCO test set}\\
\toprule
ZeroCap~\cite{ZeroCap} &\multirow{4}{*}{Zero-shot}& 2.9 & 12.0 & - &13.1\\
\textcolor{gray}{MAGIC~\cite{MAGIC}}  && \textcolor{gray}{12.9} & \textcolor{gray}{17.4} & \textcolor{gray}{-}  &\textcolor{gray}{49.3}\\
EPT\cite{ept}     && 2.2 & 12.7 & - &17.2\\
RETTA(Ours) &&\textbf{11.8}&\textbf{19.7}&\textbf{35.9}&\textbf{45.0}\\
\toprule
\end{tabular}
}

\label{mscoco_sota}
\vskip -0.1in
\vspace{-1mm}
\end{table}%

\begin{figure*}[!t]
\centering
\includegraphics[width=\linewidth]{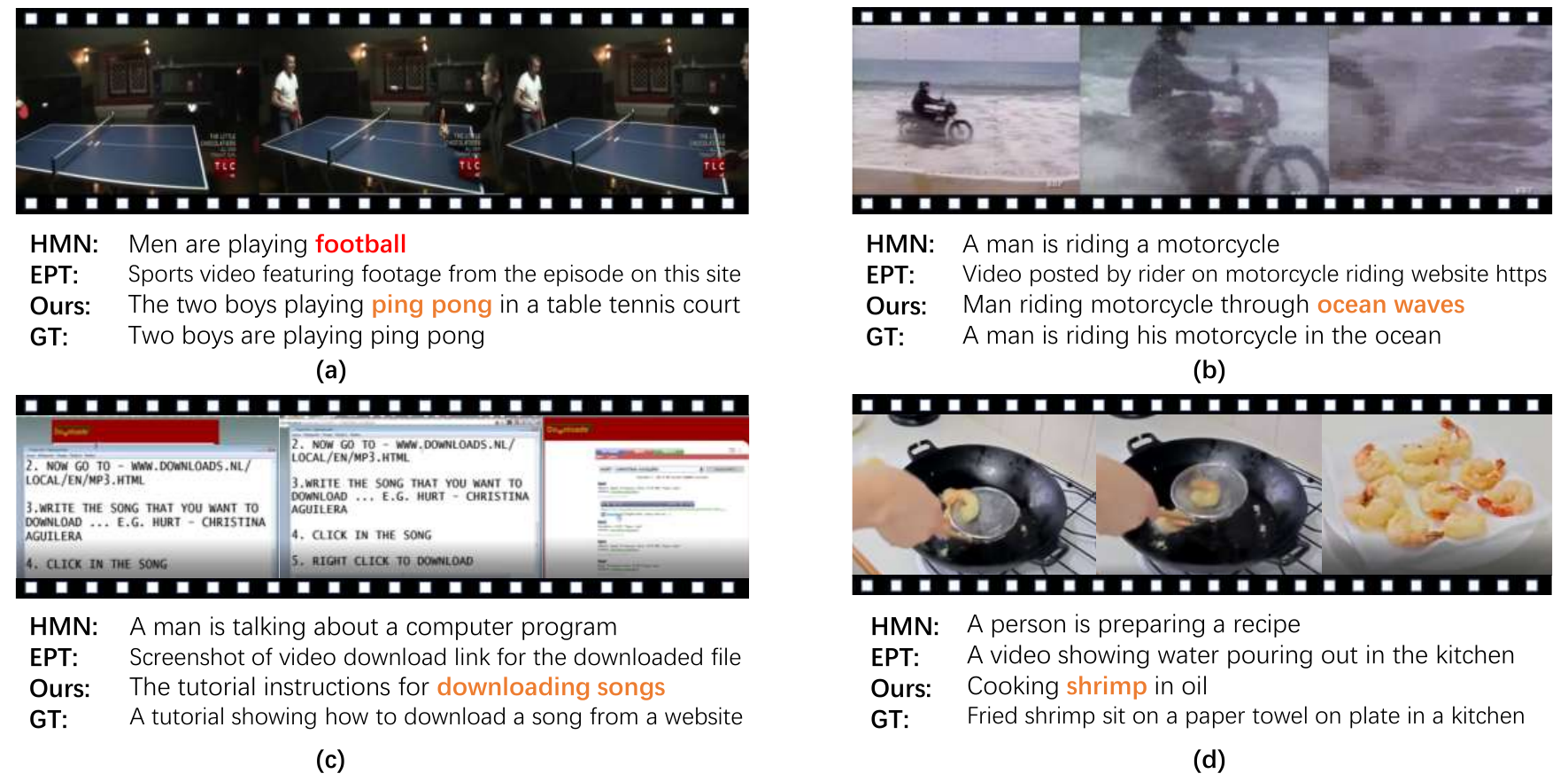}
\caption{Several video captioning qualitative examples using different methods. }
\label{fig:qualitative results}
\end{figure*}

\begin{figure*}[t]
\centering
\includegraphics[width=\linewidth]{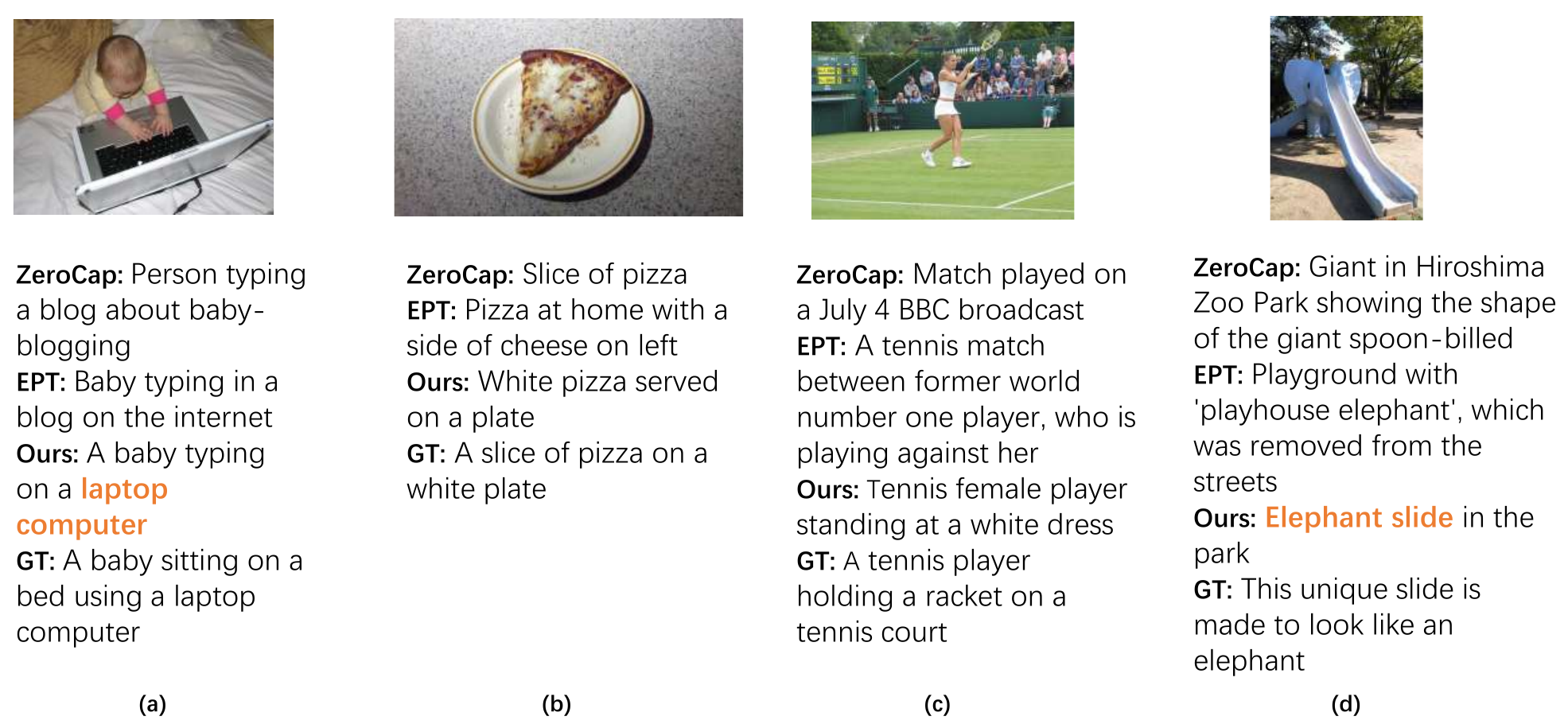}
\caption{Several image captioning qualitative examples using different methods. }
\label{fig:image captioning qualitative results}
\end{figure*}

\subsection{Qualitative Results}
Several video captioning qualitative results are shown in 
Figure~\ref{fig:qualitative results}.
We can see that the caption produced by our method is more specific than other methods.
For example, the video in Figure~\ref{fig:qualitative results} (a) shows two men playing ping pong. Our method accurately describes ``The two boys playing ping pong in a table tennis court''.
In Figure~\ref{fig:qualitative results} (b), although the fully supervised method HMN describes ``A man is riding a motorcycle'', our method more accurately identifies the ``ocean wave''.
In Figure~\ref{fig:qualitative results} (c), our method effectively describes ``The tutorial instructions for downloading songs'',  whereas the compared zero-shot method EPT merely mentions ``download'' but does not capture ``songs''.
In Figure~\ref{fig:qualitative results} (d), only our method correctly describes the video content ``Cooking shrimp in oil''. In contrast, the generated caption ``A person is preparing a recipe'' by HMN is too general. 
These 
results further show the zero-shot ability of our proposed method.

Figure~\ref{fig:image captioning qualitative results} also shows several zero-shot image captioning qualitative results.
In Figure~\ref{fig:image captioning qualitative results}~(a), our method identifies ``laptop computer'' and correctly describes the image ``A baby typing on laptop computer''.
In Figure~\ref{fig:image captioning qualitative results}~(b), while three methods all recognize ``pizza'', our method additionally captures ``plate''.
%
%
In Figure~\ref{fig:image captioning qualitative results}~(d), our method vividly generates ``Elephant slide in the park''.
\vspace{-2mm}

\section{Conclusion}
In this paper, we make the first attempt to explore the combination of a video-text retrieval model, an image-text matching model, a text alignment model, and a text generation model under the zero-shot and fast adaptation setting for video captioning.
Specifically, we use learnable tokens as an information medium connecting these frozen models. With the help of our carefully designed loss functions, these tokens are able to convey video information to the LLM in a way that serves as suitable soft prompts.
The soft target-based inference-time token update ensures fast adaptation.
%
Our proposed framework is conceptually simple and easy to implement. The extensive experiments on three benchmarks demonstrate its favorable zero-shot ability.
%
However, our work still has certain limitations. For example, it currently relies on a content-rich corpus as its database. 
\textcolor{black}{
This dependency may lead to less accurate descriptions generated by the model when the corpus is not rich in content.
Therefore, constructing a large-scale and diverse corpus is necessary to further improve the model’s descriptive capability.
}
In the future, we plan to construct a large-scale corpus to further enhance the accuracy of the model in generating visual descriptions.
\textcolor{black}{
In addition, we aim to extend our method to cross-domain video captioning tasks~\cite{crocap}, which may further enhance its generalization ability in broader cross-modal understanding scenarios.
}
We will make the code publicly available and hope our work can inspire future research in other areas.

\section{Acknowledgement}
This work was partly supported in part by the National Natural Science Foundation of China under Grants 62272438, in part by Beijing Natural Science Foundation L25700, and in part by the Fundamental Research Funds for Central Universities (E2ET1104). Yuankai Qi, Amin Beheshti, and Quan Z.
Sheng are not supported by the above-mentioned funds.


\bibliographystyle{elsarticle-num}
\bibliography{editable}

\end{document}